%% file: main.tex
\documentclass[conference]{IEEEtran}

\ifCLASSOPTIONcompsoc
  \usepackage[nocompress]{cite}
\else
  \usepackage{cite}
\fi 
\usepackage{eso-pic}

\input{headers/imports.tex}

\input{headers/math_commands.tex}
\input{headers/setup.tex} 
 

\begin{document}

\title{\vspace*{-0.5cm}%
       SoK: Data Minimization in Machine Learning%
       \vspace*{-0.25cm}
       }

\author{
\IEEEauthorblockN{%
Robin Staab$^{*\dagger}$,
Nikola Jovanovi\'c$^{*\dagger}$,
Kimberly Mai$^{\ddagger}$,
Prakhar Ganesh$^{\S}$,\\
Martin Vechev$^{\dagger}$,
Ferdinando Fioretto$^{\P}$,
Matthew Jagielski$^{\parallel}$%
}
\IEEEauthorblockA{%
$^{*}$Equal contribution \quad
$^{\dagger}$ETH Zurich, Switzerland \quad
$^{\ddagger}$University College London, UK\\
$^{\S}$McGill University / Mila, Canada \quad
$^{\P}$University of Virginia, USA \quad
$^{\parallel}$Google DeepMind, UK\\
$^{\dagger}$\{robin.staab, nikola.jovanovic, martin.vechev\}@inf.ethz.ch,\;
$^{\ddagger}$kimberly.mai@ucl.ac.uk,\;\\
$^{\S}$prakhar.ganesh@mila.quebec,\;
$^{\P}$fioretto@virginia.edu,\;
$^{\parallel}$jagielski@google.com%
}
}

\maketitle
\vspace*{-0.25cm}
\thispagestyle{plain}
\pagestyle{plain}

\AddToShipoutPictureFG*{%
  \AtPageLowerLeft{%
    \raisebox{1.9cm}{%
      \hspace*{1.75cm}%
      \parbox{\textwidth}{%
        \footnotesize This work has been accepted for publication at the IEEE Conference on Secure and
        Trustworthy Machine Learning (SaTML). The final version will be available on IEEE Xplore.%
      }%
    }%
  }%
}

\begin{abstract}
  \input{abstract.tex}
\end{abstract}
\bstctlcite{IEEEexample:BSTcontrol}

\IEEEpeerreviewmaketitle

\input{src/introduction.tex}
\input{src/regulations.tex}

\input{src/framework.tex}
\input{src/techniques.tex}
\input{src/takeaways.tex}
\input{src/conclusion.tex}

\section*{Acknowledgment}
Nikola and Robin have been supported by the SERI grant SAFEAI (Certified Safe, Fair and Robust Artificial Intelligence, contract no. MB22.00088). Views and opinions expressed are however those of the authors only and do not necessarily reflect those of the European Union or European Commission. Neither the European Union nor the European
Commission can be held responsible for them. The work has received funding from the Swiss State Secretariat for Education, Research and Innovation (SERI) (SERI-funded ERC Consolidator Grant).

Ferdinando was partially supported by NSF grants SATC-2133169, CAREER-2143706, RI-2232054 and by a Fellowship in AI Research from the LaCross Institute. The views and conclusions of this work are those of the authors only.

\bibliographystyle{IEEEtran}
\bibliography{references}

\crefalias{section}{appendix}
\appendices
\renewcommand{\thesubsection}{\alph{subsection}}
\input{appendix}

\end{document}

%% file: headers/imports.tex
\usepackage[utf8]{inputenc}                 %
\usepackage{graphicx}                       %
\usepackage[T1]{fontenc}                    %
\usepackage[hidelinks,colorlinks]{hyperref} %
\usepackage{url}                            %
\usepackage{booktabs}                       %
\usepackage{amsfonts}                       %
\usepackage{nicefrac}                       %
\usepackage{microtype}                      %
\usepackage{mathtools}
\usepackage{enumitem}
\usepackage{bbm} 
\usepackage{xspace}
\usepackage[dvipsnames]{xcolor}
\usepackage{textpos}	
\usepackage{subcaption}	
\usepackage{multirow}
\usepackage[most]{tcolorbox}
\usepackage{colortbl}
\usepackage[makeroom]{cancel}  
\usepackage{ stmaryrd }
\usepackage{wrapfig}
\usepackage{tabularx}
\usepackage{amssymb}
\usepackage{amsthm}
\usepackage{siunitx}
\usepackage{diagbox}
\usepackage[capitalize]{cleveref}
\usepackage{lipsum}
\usepackage{pifont}
\usepackage{makecell} 
\usepackage{fontawesome}   %

%% file: headers/math_commands.tex
\usepackage{amsmath,amsfonts,bm}
\usepackage{amsthm}

\theoremstyle{definition}

\def\1{\bm{1}}

\DeclareMathAlphabet{\mathsfit}{\encodingdefault}{\sfdefault}{m}{sl}
\SetMathAlphabet{\mathsfit}{bold}{\encodingdefault}{\sfdefault}{bx}{n}

\definecolor{darkblue}{rgb}{0,0,0.55}

%% file: headers/setup.tex
\newcommand{\eg}{e.g., }
\newcommand{\ie}{i.e., }

\newcommand{\smallicon}[1]{\raisebox{.1ex}{\scalebox{0.6}{#1}}}

\definecolor{pastelteal1}{HTML}{418F92}    %
\definecolor{pastelteal2}{HTML}{379FA3}    %
\definecolor{pastelbrown}{HTML}{A98274}   %
\definecolor{pastellilac}{HTML}{967BB6}   %
\definecolor{pastelcoral}{HTML}{FF746C}    %

\newcommand{\advtwire}{\textcolor{pastelteal1}{\textbf{T-Collect}}}
\newcommand{\advtpreproc}{\textcolor{pastelteal2}{\textbf{T-Preprocess}}}
\newcommand{\adviwire}{\textcolor{pastelbrown}{\textbf{I-Collect}}}
\newcommand{\advipreproc}{\textcolor{brown}{\textbf{I-Preprocess}}}
\newcommand{\advstore}{\textcolor{pastellilac}{\textbf{A-Store}}}
\newcommand{\advmodel}{\textcolor{pastelcoral}{\textbf{A-Model}}}

\newcommand{\advOne}{\textcolor{pastelteal1}{\textbf{A1}}}
\newcommand{\advTwo}{\textcolor{pastelteal2}{\textbf{A2}}}

\newcommand{\advFour}{\textcolor{brown}{\textbf{A4}}}
\newcommand{\advFive}{\textcolor{pastellilac}{\textbf{A5}}}
\newcommand{\advSix}{\textcolor{pastelcoral}{\textbf{A6}}}

\newcolumntype{x}[2]{S[table-format=#1.#2,table-auto-round]} 

\hypersetup{citecolor=blue}

\makeatletter
\AddToHook{cmd/appendix/before}{\crefalias{section}{appendix}}
\AddToHook{cmd/appendix/before}{\crefalias{subsection}{appendix}}
\AddToHook{cmd/appendix/before}{\crefalias{subsubsection}{appendix}}
\makeatother

\crefmultiformat{section}{\S#2#1#3}{\crefpairconjunction\S#2#1#3}{\crefmiddleconjunction\S#2#1#3}{\creflastconjunction\S#2#1#3}

\crefname{listing}{Lst.}{listings}
\crefname{line}{Lin.}{Lin.}
\crefname{appendix}{App.}{App.}

\newcommand{\dmml}{DMML\xspace}

\makeatletter
\renewcommand{\paragraph}{\textbf} 
    \def\@IEEEsectpunct{.\ }

    \def\para{\@startsection%
        {paragraph}%
        {4}%
        {0\parindent}%
        {0.6ex plus 0.1ex minus 0.1ex}%
        {0ex}%
        {\normalfont\normalsize\itshape\bfseries}%
        *%
    }%
\makeatother

\newcommand{\gear}[0]{\mbox{\includegraphics[height=0.25cm]{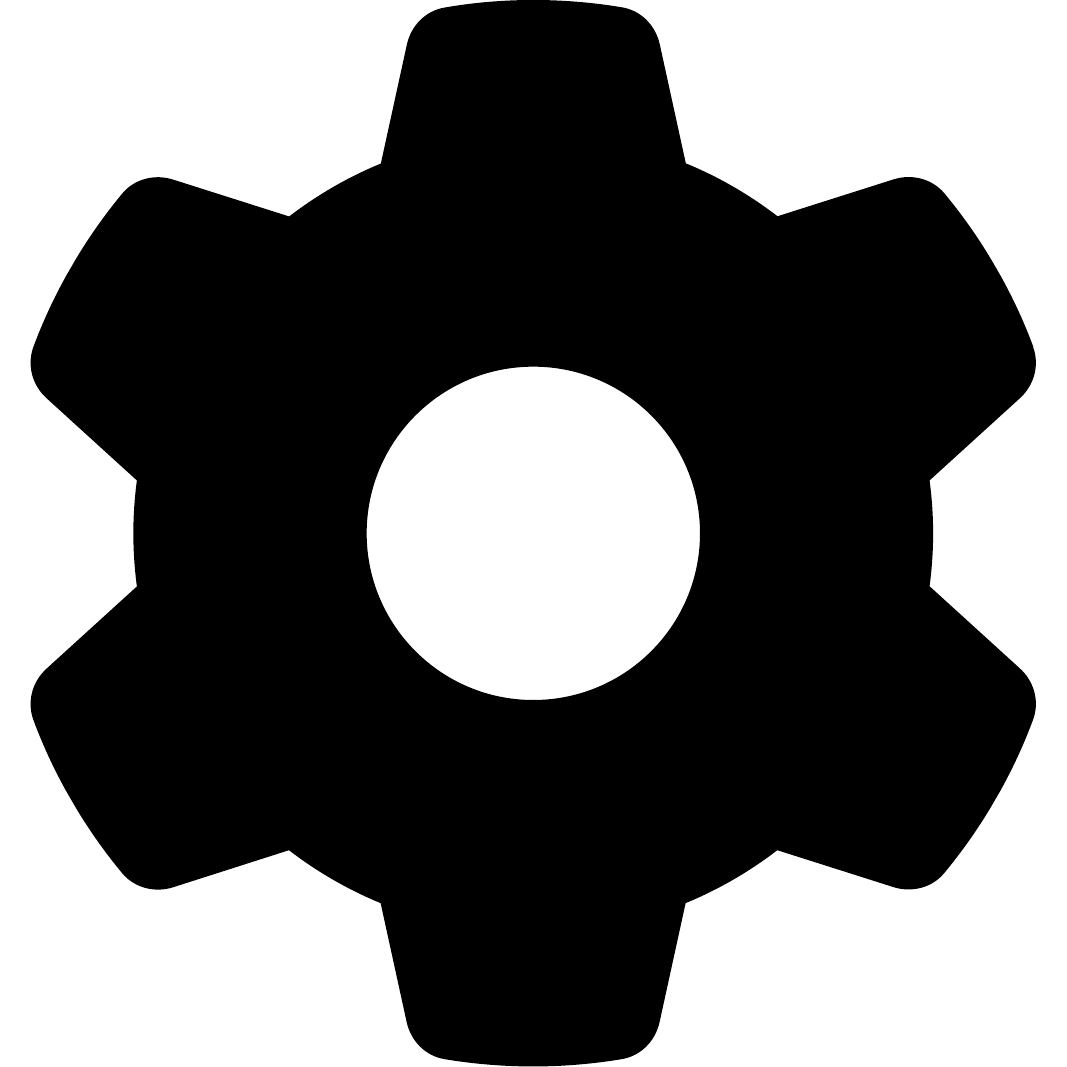}}\xspace}
\newcommand{\adv}[0]{\mbox{\includegraphics[height=0.25cm]{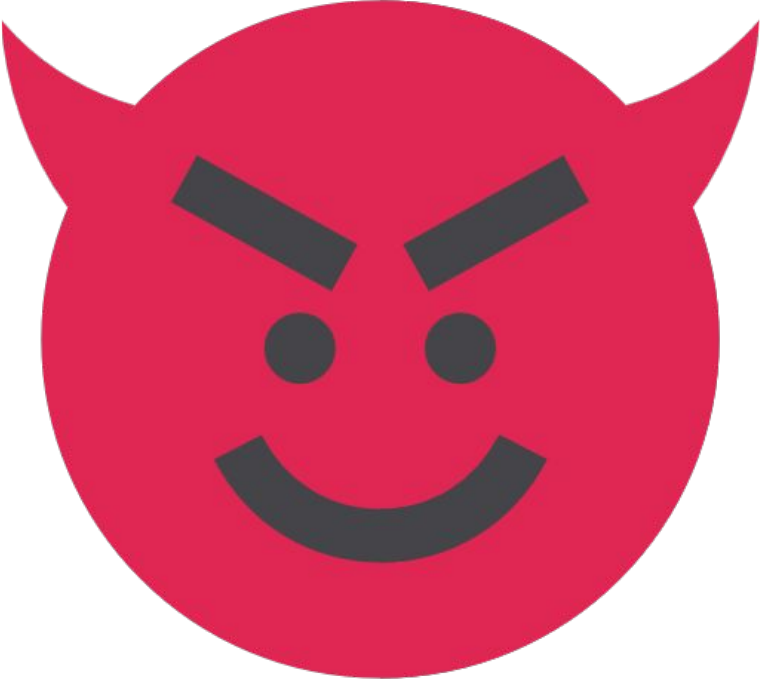}}\xspace}
\newcommand{\data}[1]{\textcolor{black}{#1}}

\PassOptionsToPackage{table,dvipsnames,x11names}{xcolor}
\usepackage{xcolor}

\newtcolorbox{quotebox}[1][]{%
  colback   = cyan!5!white,
  colframe  = cyan!60!black,
  width     = \linewidth,
  boxrule   = 0.5pt,
  arc       = 3pt,
  left      = 8pt,
  right     = 8pt,
  top       = 6pt,
  bottom    = 6pt,
  fontupper = \itshape,
  enhanced,
  breakable,
  sharp corners,
  #1
}

\newcommand{\attrib}[1]{%
  \par\hfill\normalfont\raggedleft\small---\,#1\par
}

\newtcbox{\TagBox}[1][]{
  on line,
  rounded corners,
  arc=2pt,
  colback=#1!20,          %
  colframe=#1!60!black,   %
  left=2pt, right=2pt,
  top=1pt,  bottom=1pt,
  boxsep=0pt,
  nobeforeafter,
  enlarge left by=0pt,
  enlarge right by=0pt,
  valign=center,
  fontupper=\scriptsize\sf   %
}

\newcommand{\DM}[1]{%
  \ifstrequal{#1}{H}{\TagBox[Gray]{Horizontal}}{%
  \ifstrequal{#1}{Vs}{\TagBox[BrickRed]{Vertical-S}}{%
  \ifstrequal{#1}{Vg}{\TagBox[Orange]{Vertical-G}}{%
  \ifstrequal{#1}{Vp}{\TagBox[YellowOrange]{Vertical-P}}{%
  \ifstrequal{#1}{T}{\TagBox[RoyalBlue]{Transform}}{%
     \PackageError{DM}{Undefined DM type `#1'}{}}}}}}%
}

\newcommand{\PD}[1]{%
  \ifstrequal{#1}{TC}{\TagBox[RoyalBlue]{T-Client}}{%
  \ifstrequal{#1}{TS}{\TagBox[NavyBlue]{T-Collector}}{%
  \ifstrequal{#1}{IC}{\TagBox[Cyan]{I-Client}}{%
  \ifstrequal{#1}{IS}{\TagBox[TealBlue]{I-Collector}}{%
    \PackageError{PD}{Undefined PD value `#1'}{}}}}}%
}

\newcommand{\PH}[1]{%
  \ifstrequal{#1}{Pre}{\TagBox[ForestGreen]{Pre-Hoc}}{%
  \ifstrequal{#1}{Post}{\TagBox[BrickRed]{Post-Hoc}}{%
    \PackageError{PH}{Undefined PH value `#1'}{}}}%
}

\newcommand{\PR}[1]{%
  \ifstrequal{#1}{None}{\TagBox[Gray]{None}}{%
  \ifstrequal{#1}{Exp}{\TagBox[Turquoise]{Explicit}}{%
  \ifstrequal{#1}{Imp}{\TagBox[SeaGreen]{Implicit}}{%
    \PackageError{PR}{Undefined personalization `#1'}{}}}}%
}

\newcommand{\CR}[1]{%
  \ifstrequal{#1}{Act}{\TagBox[Plum]{Act. Part.}}{%
  \ifstrequal{#1}{Tech}{\TagBox[Goldenrod]{Tech. Cap.}}{%
  \ifstrequal{#1}{None}{\TagBox[Gray]{None}}{%
    \PackageError{CR}{Undefined client requirement `#1'}{}}}}%
}

\newcommand{\AS}[1]{%
  \ifstrequal{#1}{1}{\TagBox[SkyBlue]{1 T-Collect}}{%
  \ifstrequal{#1}{2}{\TagBox[RoyalBlue]{2 T-Preprocess}}{%
  \ifstrequal{#1}{3}{\TagBox[Aquamarine]{3 I-Collect}}{%
  \ifstrequal{#1}{4}{\TagBox[Cerulean]{4 I-Preprocess}}{%
  \ifstrequal{#1}{5}{\TagBox[Orchid]{5 Store}}{%
  \ifstrequal{#1}{6}{\TagBox[Magenta]{6 Model}}{%
  \ifstrequal{#1}{NA}{\TagBox[Gray]{N/A}}{%
    \PackageError{AS}{Undefined adversarial setting `#1'}{}}}}}}}%
}}

\newcommand{\localcite}[1]{[%
  \begingroup
  \let\oldcomma=,%
  \def,{\unskip,\penalty0\hskip0pt}%
  \cite{#1}%
  \endgroup
]}

%% file: abstract.tex
Data minimization (DM) describes the principle of collecting only the data strictly necessary for a given task. It is a foundational principle across major data protection regulations like GDPR and CPRA. 
Violations of this principle have substantial real-world consequences, with regulatory actions resulting in fines reaching hundreds of millions of dollars. 
Notably, the relevance of data minimization is particularly pronounced in machine learning (ML) applications, which typically rely on large datasets, resulting in an emerging research area known as Data Minimization in Machine Learning (DMML).
At the same time, existing work on other ML privacy and security topics often addresses concerns relevant to DMML without explicitly acknowledging the connection.
This disconnect leads to confusion among practitioners, complicating their efforts to implement DM principles and interpret the terminology, metrics, and evaluation criteria used across different research communities.
To address this gap, we present the first systematization of knowledge (SoK) for DMML.
We introduce a general framework for DMML, encompassing a unified data pipeline, adversarial models, and points of minimization.
This framework allows us to systematically review data minimization literature as well as \emph{DM-adjacent} methodologies whose link to DM was often overlooked. 
Our structured overview is designed to help practitioners and researchers effectively adopt and apply DM principles in ML, by helping them identify relevant techniques and understand underlying assumptions and trade-offs through a DM-centric lens.

%% file: src/introduction.tex
\vspace{-0.5em}
\section{Introduction} \label{sec:introduction}
\vspace{-0.25em}

Systems based on machine learning are widely used across organizations and tasks, ranging from credit scoring \cite{dastile2020statistical} to fraud detection \cite{awoyemi2017credit}. 
While conventional wisdom implies that collecting more data often leads to better performance, the unchecked collection of (personal) data can violate individuals' rights while bringing minimal benefits to the actual ML model. 
To protect individuals from this, key consumer data protection regulations like the EU's General Data Protection Regulation (GDPR) \cite{gdpr} and California's Privacy Rights Act (CPRA) \cite{cpra} build on the \textit{data minimization} (DM) principle, the most explicit and referenced version is given by EU's GDPR Article 5c:

\vspace{-0.25em}
\begin{tcolorbox}[
  left=1pt, right=1pt, top=1pt, bottom=1pt,
  colback=cyan!5!white,
  colframe=darkblue!50!black,
  arc=5pt,
  boxrule=0.5pt
]
{\footnotesize\bfseries
[Personal data shall] be adequate, relevant and limited to what is necessary
in relation to the purposes for which they are processed
(``data minimisation'');
}
\vspace{-0.9em}
\attrib{\textcolor{darkblue}{\small\bfseries GDPR's Article 5(c)}}
\end{tcolorbox}
\vspace{-0.25em}

DM regulations already have a tangible impact on citizens and companies with large DM-related fines \cite{gdpr743, gdpr790} under GDPR and increasing compliance efforts \cite{gdprtracker2025Report}.
However, technical work on DM includes a broad set of disparate techniques, focusing on different objectives and methodologies for minimizing data that are difficult to compare.
Furthermore, techniques proposed in adjacent areas of ML security and privacy often have DM benefits (e.g., feature selection), but do not draw an explicit connection to DM.
This results in a critical gap where (legal) practitioners are left uncertain how common techniques, such as Federated Learning \cite{mcmahan2017communication} or Differential Privacy \cite{dwork2006calibrating}, relate to the privacy-motivated DM principle and compliance.

Even beyond legal and privacy considerations, DM techniques often provide a range of other benefits, especially when developing AI/ML-based models that commonly require lots of data.
Here, {careful data minimization can often improve model performance \cite{vergara2014review} and reduce computational and storage costs during algorithm development \cite{nalepa2019selecting}}. 
Importantly, the pursuit of \textit{minimizing} data collection not from privacy but from a performance, cost, or effort perspective has a long history in machine learning research. This results in a wide range of works \cite{shorten2019survey,gao2023retrieval,zhu2024survey,settles2009active,lei2023comprehensive,Fung2010} which implicitly provide data minimization without explicitly recognizing it, such as active learning or data curation \cite{li2024survey}.
Given that disconnect, our work explicitly focuses on the following two questions:

\vspace*{-0.5em}
\begin{center}
\textcolor{darkblue}{\textit{What defines data minimization in machine learning? Which existing techniques satisfy some form of data minimization?}}
\end{center}
\vspace*{-0.5em}

We address these questions by, for the first time, systematizing existing work on data minimization and adjacent methods that achieve \emph{Data Minimization in Machine Learning} (\dmml).
Our goal is not to give use-case dependent recommendations but to provide a structured view (\cref{fig:overview}) of disconnected areas that implicitly study aspects of DM and introduce a comprehensive resource that practitioners and researchers can refer to when applying or developing DM methods.
While some prior works propose techniques explicitly motivated by DMML (we cover those in~\cref{subsec:techniques:explicit}), to the best of our knowledge, there were no prior attempts to systematize the area or provide a DM-centric perspective on relevant technical work.
\input{figures/overview.tex}

\para{$\blacksquare$ Key Contributions}%
We make the following contributions:
\begin{enumerate}[leftmargin=*]
    \item Drawing from the regulatory background and key motivations behind DM (\cref{sec:regulations}), we introduce a general framework for \dmml, defining actors, data pipelines, privacy risks, and dimensions of \dmml (\cref{sec:framework}).
    \item Using our framework we systematically analyze DM-adjacent techniques, for the first time highlighting their relationships to DM in a structured way (\cref{sec:techniques}).
    \item We highlight key takeaways and open research questions on the topics of DM operationalization, classification, and standardized evaluation (\cref{sec:takeaways}).
    \item We provide a comprehensive overview of the datasets used in work on DM-adjacent techniques (\cref{sec:datasets}).
\end{enumerate}

%% file: figures/overview.tex
\begin{figure*}
    \centering
      \includegraphics[trim={0.5cm 1cm 0.1cm 3cm},clip,width=\linewidth]{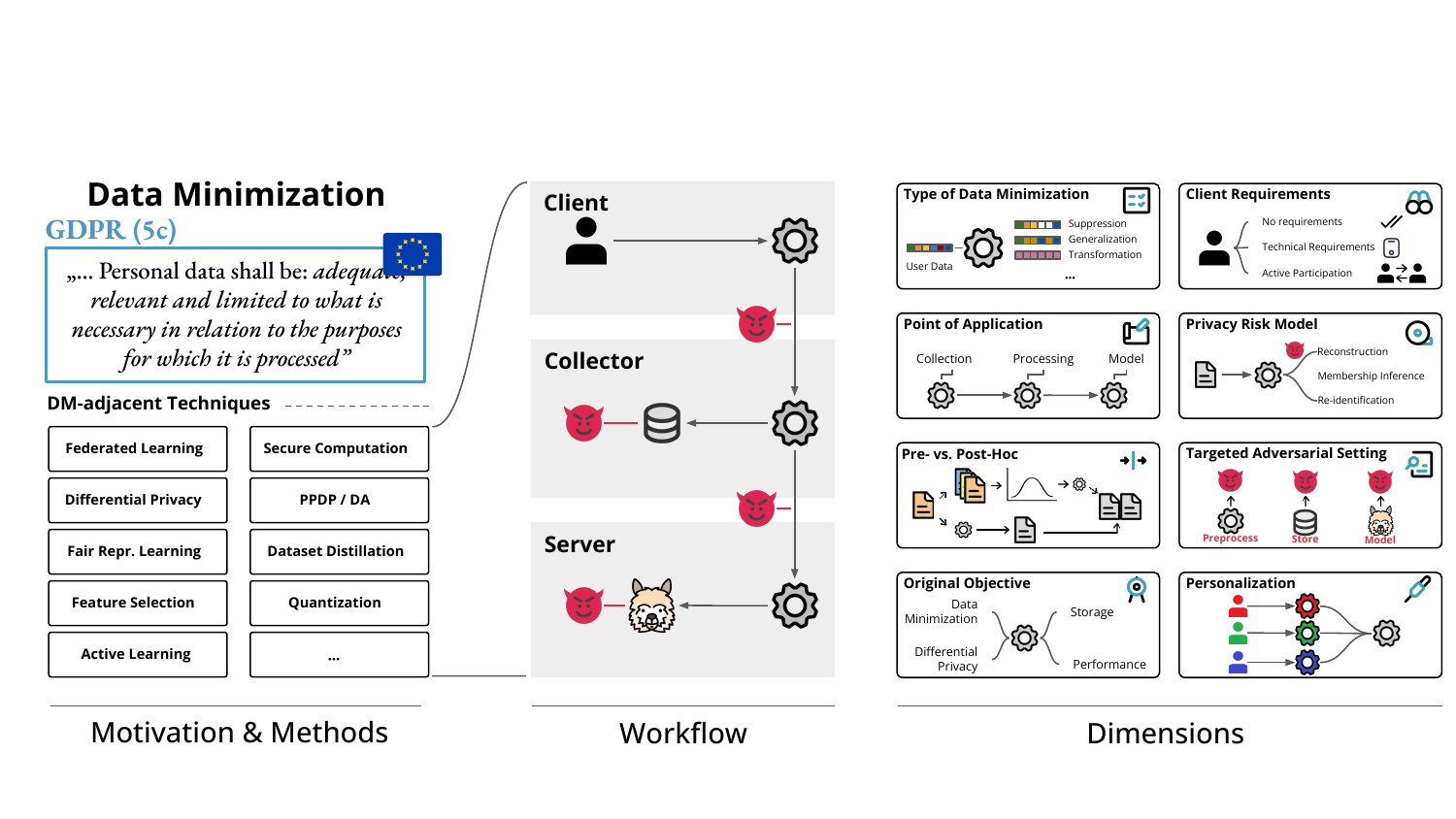}
      \caption{Overview of our work. The Data Minimization (DM) principle is referenced by regulations such as GDPR (\emph{top left}, \cref{sec:regulations}). At the same time, DM as the \emph{act of minimizing data} is present in some form as part of various \emph{DM-adjacent techniques}, without this connection being made explicit, which leads to confusion (\emph{bottom left}). We unify these under a joint \dmml framework (\cref{sec:framework}). In particular, a well-defined \dmml workflow (\emph{middle}, detailed in~\cref{fig:pipeline}) and a list of relevant dimensions of DM techniques (\emph{right}) allow us to systematically analyze all relevant techniques (\cref{sec:techniques}).}
      \label{fig:overview}
      \vspace{-1em}
\end{figure*}

%% file: src/regulations.tex
\section{Data Minimization Regulations} \label{sec:regulations}
This section reviews the existing data minimization regulations as well as their instantiation in \dmml. 

\para{Regulations}
Data minimization is a core principle in several data protection laws, including GDPR (EU) \cite{gdpr}, CPRA (California) \cite{cpra}, and LGPD (Brazil) \cite{lgpd}, generally requiring personal data to be \textit{``adequate, relevant, and limited to what is necessary``} for specific processing purposes (Article 5C, GDPR). Similarly, the CPRA stipulates that \textit{``a business' collection, use, retention, and sharing of a consumer's personal information shall be reasonably necessary and proportionate to achieve the purposes for which the personal information was collected or processed, [...] and not further processed in a manner that is incompatible with those purposes''} (California Civil Code §1798.100(c)). CPRA also encourages organizations to assess the need for new data processing activities, apply safeguards for personal data, and reduce the use of sensitive personal information where feasible. Crucially, these regulations define \emph{personal data} broadly, encompassing any data linkable directly or indirectly to an identifiable individual \cite{gdpr}, going beyond categories like personally identifiable information (PII) such as names or social security numbers. Specifically, GDPR's definition of personal data extends to more subtle identifiers, including, \eg IP addresses, cookies, and online behavioral data.

Recent regulatory attention has highlighted the importance of adhering to data minimization principles. For instance, scrutiny of facial recognition technologies and AI chatbots has intensified in response to concerns about large-scale data collection practices and their alignment with data minimization principles \cite{clearviewaustria, clearviewdutch, clearviewfrance, clearviewgreece, clearviewitaly, replika}. Moreover, the rise of large-scale AI and Large Language Models (LLMs) has generated tension around data minimization's practicality, as critics question whether competitive performance can coexist with minimized personal data processing \cite{icouk}. These challenges have highlighted the need for more technically grounded guidance on how to implement minimization in AI contexts \cite{edpb2024ai, icouk2024genai}.

\para{The \dmml Gap}
In practice, the abstract nature of regulatory definitions combined with technical constraints often leads to narrow interpretations of data minimization. Many privacy-focused studies emphasize removing explicitly sensitive data or PII, while overlooking less obvious or publicly available data that may still fall under regulatory scope \cite{solove2005taxonomy}. 

Interestingly, many popular ML methodologies (e.g., feature selection, federated learning, and differential privacy) implicitly embody elements of data minimization. Yet, these techniques rarely frame their objectives in terms of regulatory compliance or DM principles, and even methods that explicitly focus on DM often lack clarity on which aspects of DM they address and how they relate to the DMML pipeline (see \cref{ssec:framework:pipeline}).
Consequently, ML practitioners lack a broader overview of DMML-related techniques, increasing their uncertainty whether established methods fulfill legal DM obligations. This raises the question: can we provide \emph{clearer alignment and interpretability between technical and regulatory definitions?}

%% file: src/framework.tex
\vspace{-0.5em}
\section{The \dmml Framework} \label{sec:framework}
\vspace{-0.5em}

To address this question, our work proposes a structured and unified framework explicitly tailored for \emph{Data Minimization in Machine Learning} (\dmml).
For this, we first introduce standardized terminology (including involved actors and key roles) designed to reconcile differences across the ML and regulatory communities (\cref{ssec:framework:actors}).
Then, we provide an overview of a high-level \dmml pipeline that identifies key points of data transformation and potential adversarial interception (\cref{ssec:framework:pipeline}).
Next, we discuss common metrics and criteria used to quantify the effectiveness of DM techniques (\cref{ssec:framework:quantifying}).
Lastly, we outline a set of \emph{dimensions} that characterize and differentiate \dmml methods (\cref{ssec:framework:dimensions}).

\input{figures/pipeline.tex}
\subsection{\dmml Actors} \label{ssec:framework:actors}

A \dmml pipeline (shown in \cref{fig:pipeline}, introduced in detail in \cref{ssec:framework:pipeline}) involves the following actors: 

\begin{enumerate}[leftmargin=*,parsep=0pt, topsep=4pt]

\item \emph{Data Owner} (\textbf{Client}): refers to every party that contributes (their) data at \emph{any} step of the \dmml pipeline (\cref{ssec:framework:pipeline}). 
In scenarios where contributed data is sensitive (\cref{sec:regulations}), the Client is exposed to privacy risks. When public data (\cref{app:data}) suffices, we consider no specific Client.

\item \emph{Data Collector} (\textbf{Collector}): refers to every party collecting data from Clients for later processing.

\item \emph{Service Provider} (\textbf{Server}): an actor that trains machine learning model(s) on data they received from the Collector and later performs inference on newly collected data. 
In some scenarios, the Collector directly owns the models, \ie the Server and the Collector are the same party. In this case, we call the unified party the Collector. 
\end{enumerate} 
We generally assume that Collector and Server do not change between model training and inference stages.
 
\para{Running Example} To illustrate the roles of the actors, consider the following example that we will refer to throughout this section. 
A hospital (Collector) collects data about their patients (Clients) in various ways: paper forms, online surveys, and medical devices.
The hospital's goal is to train a ML model on all collected data to predict each patient's likelihood of developing a certain disease. As they do not have the resources to train the model and run inference themselves, they offload both tasks to a cloud service provider (Server).

\subsection{\dmml Pipeline} \label{ssec:framework:pipeline}
Next, we describe the \dmml pipeline (illustrated in~\cref{fig:pipeline}) following the flow of data and the transformations applied to it by different actors.  
Therein, each \gear denotes a potential data transformation; note that between any two transformations, an adversary (\adv) could intercept the transformed data (which we describe in detail in \cref{ssec:framework:quantifying}).

During the training stage (top of~\cref{fig:pipeline}), the Collector, \eg the hospital in our example, aggregates \data{training data} such as laboratory results, demographic fields, and device readings from multiple Clients (patients). Each Client may first apply a local transformation, such as a withholding attributes or noise addition, before disclosure (\gear). 
The Collector may then apply an additional preprocessing step (\gear), such as merging or dropping some measurements and removing direct identifiers, to obtain \data{preprocessed training data}. 
The Server (\emph{right}) then trains a machine learning model on this \data{preprocessed training data}, where the model training itself can be seen as another data transformation (\gear) whose output is the set model parameters used for later inference.
In both training and inferences stages, the Collector may retain a further transformed (\gear) representation of the data as \data{stored data}.
This can aid future model improvements or be useful for a different ML pipeline (\eg enabling faster responses).

The inference stage (bottom of~\cref{fig:pipeline}) mirrors the flow above, assuming a black box scenario where the trained model is only accessible to Clients through the Collector.
New clinical measurements, potentially already preprocessed on the patient's device (\gear) are submitted as \data{inference data}.
The Collector (hospital), again executes its preprocessing routine (\gear), yielding \data{preprocessed inference data} that the cloud provider consumes to generate a response from the trained model, \eg a disease-risk score. 
The model prediction is returned to the hospital, and, optionally, back to the Client. 

\subsection{Quantifying \dmml} \label{ssec:framework:quantifying}
Equipped with the overview above, \emph{data minimization in an ML context} is defined as \textcolor{darkblue}{\emph{any data modification within the pipeline that reduces the amount of information present from the perspective of any of the adversaries}} (illustrated in \cref{fig:pipeline} as \gear and \adv respectively and introduced shortly).

\para{Baseline Metrics} One can measure the impact of a data minimization intervention with a wide-range of metrics, corresponding to different motivations discussed in~\cref{sec:introduction}. 
\textbf{\textit{Utility}} quantifies the impact of data minimization on the performance of the model trained on minimized data, and is often measured as the accuracy on some downstream task (e.g., predicting a specific disease). 
Other options here include the absolute/relative \emph{drop in accuracy} compared to a reference model~\cite{goldsteen}, specific probabilistic measures~\cite{tran23}, or other regulation-guided metrics \cite{anciaux2024new}. 
The impact of data minimization on \textit{size} is commonly either directly measured by the amount of data that is \emph{passed/stored}, or via relevant downstream proxies such as \emph{latency/throughput} at different parts of the pipeline which depend on data size.
In contrast to privacy, all data is considered equal from this perspective.
There may be other objectives more loosely tied with DM, \eg {\textbf{\textit{fairness}}, which is a core principle in GDPR~\cite{gdpr} and can be measured in terms of disparate impact metrics~\cite{hardt2016eq}.

It is important to note that (in the limit) the objectives of privacy and utility are fundamentally at odds, implying a \emph{utility-privacy tradeoff} \cite{staab24vdm}. While regulations (\cref{sec:regulations}) generally have strict privacy requirements (``limited to what is necessary''), in practice, this tradeoff is more continuous. As we show below and expand in \cref{app:adversarial}, the adversarial risk is thereby commonly evaluated w.r.t. the empirical utility of the minimized data, trying to achieve an ``optimal'' tradeoff.

\para{Adversarial Setting} We view privacy through an \emph{Adversary} that can access the data at different points in the pipeline either directly (\eg untrusted Collector) or via a \emph{breach}.
We distinguish six adversaries based on their data access.
\begin{enumerate}[label=A\arabic*),ref=A\arabic*,leftmargin=*]
    \item \emph{Adv.~\advtwire}: Gains access to the raw training data during the training data collection step. 
    \item \emph{Adv.~\advtpreproc}: Gains access to the training records, \emph{after} preprocessing, but \emph{before} the model training.
    \item \emph{Adv.~\adviwire}: Intercepts the inference data as it is submitted during the inference data collection step.
    \item \emph{Adv.~\advipreproc}: Gains access to the preprocessed inference data, just before it is sent to the cloud for prediction in the inference step.
    \item \emph{Adv.~\advstore}: Gains access to the stored data at any point by accessing the Collector’s storage container.
    \item \emph{Adv.~\advmodel}: Gains access to the model any time after the training step.
\end{enumerate}
Note that their exact location (\eg Adv. \advOne{} being on the wire, being the Collector itself, or on the Client side) does not change the privacy risk analysis and is thus abstracted away in our framework. 
At the same time, as detailed in \cref{sec:techniques}, the stage of the DDML pipeline at which an adversary gains access induces a partial order on threat models: a defense that thwarts an early adversary such as \advtwire{}, where the data collector itself is untrusted, in many cases protects against later adversaries, especially if unminimized data never leaves the user, whereas a defense designed for a later adversary, like \advmodel{}, assumes that every preceding party is trusted.

\para{Measuring Privacy Risk} For adversaries with data access (Adv. \advOne{}--\advFive{}) we can measure the privacy risk as follows:
\begin{itemize}[leftmargin=*]
    \item The number of (sensitive) features accessed in total or per Client. This is a na{\"i}ve measure as it ignores the true privacy risk, \eg correlations between features~\cite{aware}. 
    \item The success of inference/reconstruction attacks, which approximate the highest achievable accuracy in predicting original from minimized data. While more indicative of the true privacy risk, there are several open questions: 
    {\it (i)} How good is the approximation of the optimal attack? 
    {\it (ii)} Does worst-case or the population-level average risk matter more? 
    {\it (iii)} When is reconstruction considered successful? 
    \item The success of linkability/reidentification attacks~\cite{dpwp}, which link the minimized data to other data sources.
    \item The success of singling out attacks~\cite{dpwp}, which aim to identify a particular individual in the breached data.
\end{itemize}

We provide a detailed overview of these metrics, including their use cases and potential pitfalls, in \cref{app:adversarial}.
While \advOne{}–\advFive{}{} represent direct data exposure, \advSix{} represents indirect data leakage through the learned parameters and we can measure the indirect privacy risk to the training data via:
\begin{itemize}[leftmargin=*]
    \item Membership inference attacks~\cite{Shokri2017}, which aim to determine whether a data point was used for model training.
    \item Model inversion~\cite{ModelInv} and attribute inference attacks~\cite{attrinf}, which aim to recover \emph{representative} training inputs or parts of actual training inputs, respectively.
    \item Data extraction attacks~\cite{carlini2023quantifying}, which aim to recover whole training inputs, amendable to further inference attacks. 
\end{itemize}
\vspace{-0.25em}
\subsection{Dimensions of \dmml} \label{ssec:framework:dimensions}  
\vspace{-0.25em}

\input{src/dimensions_2}

\para{Other Dimensions}
The above dimensions, while enabling the categorization and comparison of \dmml(-adjacent) techniques, are not exhaustive.
Additional dimensions not deemed directly relevant from the DM perspective may include applicable modalities (\eg text, audio, tabular) or the intended application (\eg recommendation, content generation, classification), which we overview in \cref{sec:datasets}. 
Further, we implictely assume DM at the beginning of a system lifecycle, treating later \emph{corrective DM} as failed attempts, as data may already been exposed to privacy risks.

%% file: figures/pipeline.tex
\begin{figure*}[t]
    \centering
      \includegraphics[trim={0.5cm 1.2cm 3cm 3cm},clip,width=0.88\linewidth]{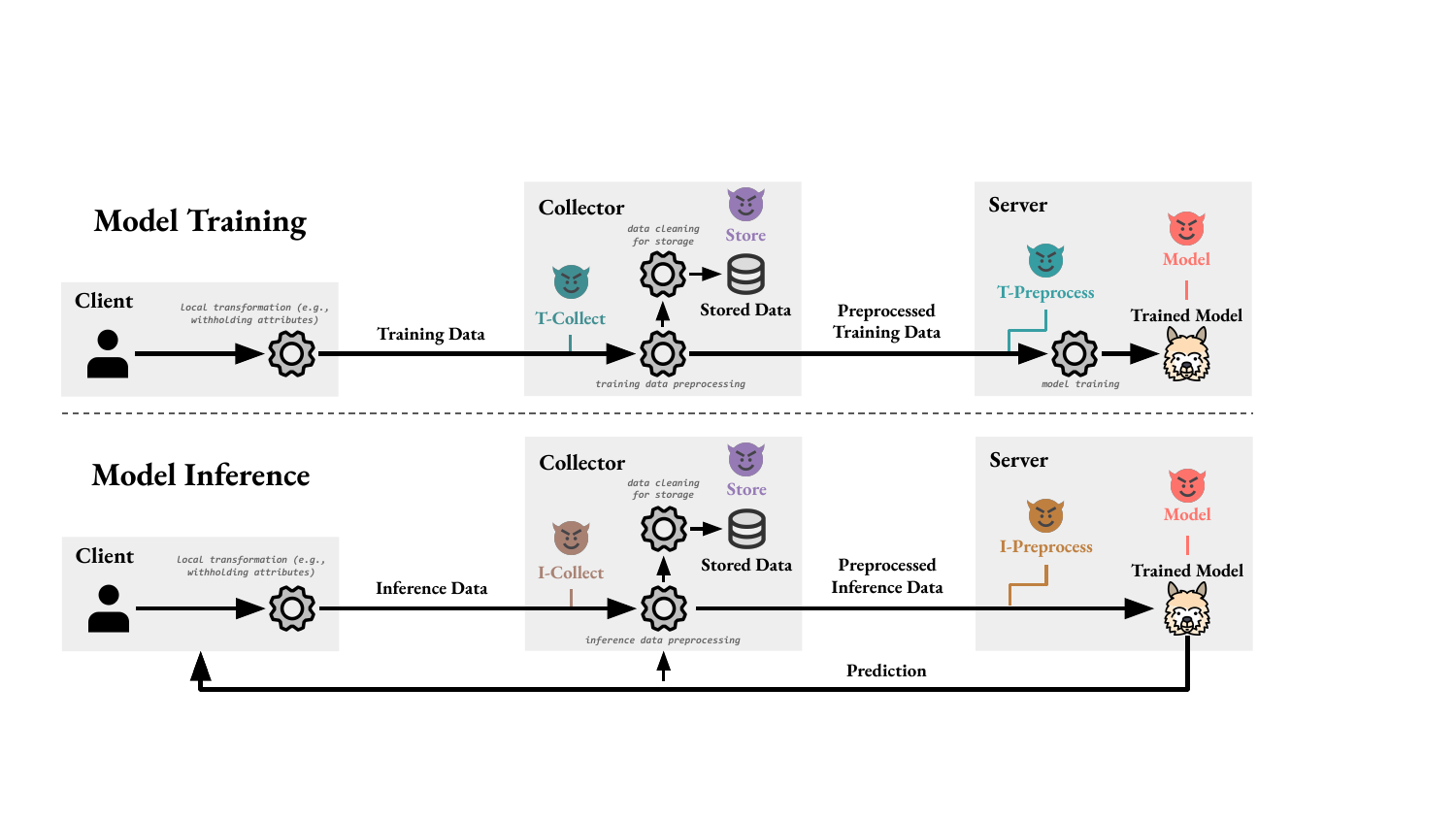}
      \caption{Illustration of \dmml actors (\cref{ssec:framework:actors}), pipeline (\cref{ssec:framework:pipeline}), and adversaries (\cref{ssec:framework:quantifying}). During training and inference, data is provided by clients which may transform it (\gear) before sending it to the collector. The collector can further transform the data and may store it for later use. Finally, data is sent to a server to train the final model (another transformation). Between any two transformations an adversary (\adv) could intercept the data, threatening client's privacy.}
      \label{fig:pipeline}
      \vspace{-1em}
\end{figure*}

%% file: src/dimensions_2.tex
Next, we introduce a set of \emph{dimensions} that characterize \dmml methods. 
They correspond to the columns of our overview table (\cref{tab:overview_adjacent}), summarizing \dmml-adjacent techniques from \cref{sec:techniques}. 
Importantly, throughout this and the following sections, we will refer to techniques in their isolated and canonical instantiations. 
While techniques are commonly amenable to combinations, their respective properties across our dimensions combine naturally. Keeping them separate allows for clearer discussions of individual benefits and drawbacks without having to describe all combinations.
To ground the discussion we revisit our running example: a hospital (Collector) gathers patient data (Clients) for disease prediction, outsourcing model training and inference to a cloud provider (Server).

\para{Original Objective}
What is the \emph{intended} goal of the method, and how does it relate specifically to data minimization?
For some methods, DM emerges only as a side effect of another primary objective.
For instance, a hospital might use dataset distillation (\cref{subsec:techniques:dd}) to reduce the burden of training data storage. While their primary aim is efficiency, this simultaneously achieves a degree of DM by significantly compressing data before sending it to the Server.

\para{Type of DM}
What specific mechanism underlies the data minimization? We differentiate between example removal (\emph{horizontal}, e.g., collecting fewer patient records), reduction of information within each data point (\emph{vertical} via (i) not collecting certain patient attributes (\emph{suppression}), (ii) collecting, e.g., only the age ranges instead of exact ages (\emph{generalization}), or (iii) \emph{perturbing} certain attributes), and non-interpretable transformations (\emph{transformative}, e.g., encoding patient data as compact embeddings (\cref{subsec:techniques:frl})).
Each type affects privacy risks differently; horizontal DM reduces the risk at a population level but leaves individual privacy exposed. Certain techniques blend these mechanisms, for instance, active feature-value acquisition (\cref{subsec:techniques:act_learn}).

\para{Points of DM Application}
At which stage (\eg \emph{Training} or \emph{Inference}) and by whom (\emph{Client} or \emph{Collector}) is minimization executed?
As shown in \cref{fig:pipeline}, DM techniques can be applied across the entire training/inference pipeline. 
Importantly, the point of application encodes common trust assumptions: Client-based techniques assume limited or no trust in the Collector, whereas Collector- or Server-side DM presumes that upstream parties are trusted to access certain full-resolution data.
Several explicit DM (\cref{subsec:techniques:explicit}), as well as DM-adjacent, techniques (e.g., federated learning, \cref{subsec:techniques:federated}) apply DM directly on the Client side. Though not generally applicable (for instance, federated learning only applies to training data), these methods aim to ensure that patient data is never collected in a non-minimized form. 
In contrast, widely studied \emph{privacy-preserving data publishing} techniques (\cref{subsec:techniques:ppdp}) or even \emph{synthetic data generation} (\cref{subsec:techniques:syn_data}) require a baseline set of full-resolution in-domain patient data before minimized versions can be produced (implicitly putting trust in the respective Collector).
Lastly, techniques aiming to protect from \advmodel{}, such as central differential privacy (\cref{subsec:techniques:dp}), only get applied on the Server, assuming all parties are trusted.
This dimension closely reflects the agency of Clients: whether empowered to withhold irrelevant personal data or relying on the hospital's/server's decisions about minimization.

\para{Pre- or Post-Hoc}
Is minimization performed before (\emph{Pre-Hoc}) or after (\emph{Post-Hoc}) data leaves the Client?
For instance, in FL (\cref{subsec:techniques:federated}), minimization occurs directly on patients' devices, ensuring no full-resolution data leaves the Client (Pre-Hoc).
On the contrary, Post-Hoc techniques, such as privacy-preserving data publishing (\cref{subsec:techniques:ppdp}) or central DP (\cref{subsec:techniques:dp}), first require collecting full user data (e.g., comprehensive patient records), only to minimize it later upon data release (training).
While this dimension overlaps with the point of DM application, it captures an important distinction in trust and accountability: whether clients retain agency over minimization or rely on a trusted collector to enforce it.

\para{Client Requirements}
What are the practical requirements each technique places upon Clients?
We define three categories of technical requirements: Several explicit and DM-adjacent techniques like feature selection (\cref{subsec:techniques:f_sel}) have \emph{no requirements}, enabling even analog data collection (e.g., paper forms).
The category \emph{technical capability} assumes that Clients have a local device that enables them to perform more involved data transformations. While we do not directly differentiate the level of technical capability required (e.g., local DP only adds noise while federated learning computes gradient), we do differentiate between methods that can be performed once in an offline fashion versus ones that require \emph{active participation} of Client's in the actual computation (\eg in secure multi-party computation (\cref{subsec:techniques:sec_comp}) or federated learning (\cref{subsec:techniques:federated})).
Identifying these requirements explicitly is essential to gauge realistic applicability; overly restrictive requirements may render certain minimization approaches impractical despite theoretical advantages.

\para{Personalization}
Does the method incorporate personalization (explicitly or implicitly), or are all Clients treated uniformly?
Returning to our example, personalized federated learning~\cite{personalizedfl} could be used to explicitly tailor training and DM to each patient's characteristics, potentially improving predictive accuracy and patient acceptance---at the cost of higher technical requirements and lower privacy protection.

\para{Targeted Adversarial Setting}
For privacy-oriented methods, which specific adversaries (as described in \cref{ssec:framework:quantifying}) are they primarily defending against?
For instance, if the hospital adopts a technique specifically targeting Server-side adversaries (i.e., \advTwo{},\advFour{},\advSix{}), this implicitly indicates trust towards the hospital (Collector). Although the hospital's primary privacy objectives may be explicit, many DM-adjacent methods inadvertently provide secondary protections against additional adversaries beyond their intended scope. Likewise more efficiency focused techniques operate under no targeted adversarial scenario, denoted by N/A.

\para{Privacy Risk Model and Guarantees}
What privacy risk does the method explicitly address, and what formal guarantees (if any) does it provide?
Methods providing statistical guarantees, such as DP (\cref{subsec:techniques:dp}), are intuitively preferred in regulatory compliance. At the same time, DP's membership-inference $(\epsilon, \delta)$-guarantees do not translate to other adversaries, such as attribute reconstruction (i.e., inferring patient attributes). Here, methods lacking explicit guarantees might nonetheless offer stronger privacy protection, emphasizing the importance of aligning expected risks and guarantees.

%% file: src/techniques.tex
\section{Analysis of DM-adjacent Techniques} \label{sec:techniques}
\input{tables/techniques.tex}
Next, we apply the \dmml framework introduced in~\cref{sec:framework} to systematically analyze techniques that relate to data minimization in a machine learning setting, highlighting connections that previously may not have been explicitly recognized. 
In each of~\cref{subsec:techniques:federated,subsec:techniques:dp,subsec:techniques:sec_comp,subsec:techniques:ppdp,subsec:techniques:frl,subsec:techniques:syn_data,subsec:techniques:dd,subsec:techniques:ds,subsec:techniques:d_aug,subsec:techniques:f_sel,subsec:techniques:act_learn,subsec:techniques:rag,subsec:techniques:model_comp} we introduce one \emph{Technique} (\eg Federated Learning), identify representative \emph{Instantiations} (\eg FL model training), and discuss their \emph{relationship to DM}.
Where applicable, we list additional \textbf{DM observations}, \ie the unique ways the particular technique interacts with parts of our DM framework in certain settings, and highlight research gaps that can inspire future work (marked with {\faLightbulbO}).
Lastly, we move away from \textit{DM-adjacent} techniques and, in~\cref{subsec:techniques:explicit} introduce and discuss recent research that \emph{explicitly} targets \dmml.
Throughout this, we explicitly aim for breadth rather than depth, \ie our goal is to provide an overview of all techniques whose common instantiations are in some way relevant to DMML. 
We do not attempt to cover all variants of these techniques or their combinations, nor to provide an exhaustive technical overview.
Our review is unopinionated: we do not aim to rank techniques or provide recommendations, as which method is most suitable greatly varies across use cases.

\subsection{Federated Learning}
\label{subsec:techniques:federated}
Federated learning (FL)~\cite{mcmahan2017communication} is an approach to distributed machine learning in which a collector, instead of collecting data into a central database to train a model, collects only model updates from users. This requires Clients to have the capability to locally update models. Further, FL typically assumes the collector and server to be the same party, as aggregation involves applying users' model updates.

\para{Relationship to DM}
Federated learning prevents the collector from directly inspecting training data, providing an intuitive protection against adversaries \advtwire{} and \advstore{}. However, it has been shown that this intuitive protection is leaky: a curious collector can \emph{extract} full training examples from a user's model update \cite{zhu2019deep, boenisch2023curious}. As a result, to provide provable guarantees against \advtwire{} and \advstore{}, federated learning is often combined with lightweight secure computation and differential privacy (see Sections \ref{subsec:techniques:dp} and \ref{subsec:techniques:sec_comp}). These additions also extend protection to \advtpreproc{} and \advmodel{} \cite{bonawitz2016practical, kairouz2021advances}. We note that federated learning requires clients to update models locally, necessitating both client-side computation and participation in a protocol.

\subsection{Differential Privacy}
\label{subsec:techniques:dp}
Differential privacy (DP)\cite{dwork2006calibrating} is a formal guarantee of privacy provided by carefully designed randomized algorithms. This guarantee bounds the impact that any training example can have on the output of a data analysis process (such as training a machine learning model). Formally, an algorithm $A$ satisfies $(\varepsilon, \delta)$-differential privacy if, for any two datasets $X, X'$ differing in only one row and any set of outputs $O$,
\[
\Pr[A(X)\in O] \le e^{\varepsilon}\Pr[A(X')\in O] + \delta.
\]

\para{Relationship to DM}
Differential privacy can be instantiated in multiple ways, to hide information from different actors. Central differential privacy \cite{dwork2006calibrating} permits a data aggregator to collect all data in the clear before training a model, providing protection only from adversary \advmodel. It provides no protections from other adversaries, but can be combined with techniques such as federated learning (\cref{subsec:techniques:federated}) or secure computation (\cref{subsec:techniques:sec_comp}) to broaden its protections.

It is possible to provide differential privacy guarantees earlier in the data collection process, in both local differential privacy \cite{kasiviswanathan2011can} and distributed differential privacy \cite{dwork2006calibrating, bittau2017prochlo}. Instead of collecting data ``in the clear,'' algorithms satisfying these guarantees allow the client to randomize their training data to protect against adversaries \advtwire{}, \advtpreproc{}, and \advstore{} (on top of \advmodel). Typically, local DP leads to heavy performance degradation \cite{kasiviswanathan2011can}, while distributed differential privacy requires additional infrastructure.

\smallskip\noindent\paragraph{DM Observations:}
\begin{itemize}
    \item For a given level of model accuracy, differential privacy requires collecting more data \cite{dwork2006calibrating, dwork2015robust, tramer2020differentially}. However, in central DP, this is at odds with horizontal DM, producing a tradeoff between Adversary \advmodel~ and Adversaries \advtwire/\advtpreproc/\advstore.
    \item Differentially private training benefits from reducing irrelevant features \cite{tramer2020differentially}. As a result, improving vertical DM can simultaneously improve DP training.
    \item Local DP can be applied at inference time \cite{du2023dp}, although with an upper bound on utility \cite{carlini2021private}.
\end{itemize}

\subsection{Secure Computation}
\label{subsec:techniques:sec_comp}
Secure computation aims to enable model training and inference on data whose privacy is protected via cryptographic techniques.
The most prominent approaches use \emph{Fully Homomorphic Encryption} (FHE)~\cite{Gentry09} which allows computations to be performed directly on encrypted data.
This has been instantiated successfully for both model inference~\cite{GiladDLLNW16,BadawiJLMJTNAC21,BrutzkusGE19,LeeKLCEDLLYKN21,LeeLLKKNC21,LuHHMQ20} and training~\cite{kim2018secure,kim2018logistic,nandakumar19towards,lou2020glyph,mihara2020neural}, where the latter is still limited to simple models (\eg logistic regression, shallow fully-connected networks) due to the challenge of efficiently evaluating complex models under FHE.
In another attempt to overcome those challenges, some combine FHE with \emph{Secure Multi-Party Computation} (MPC)~\cite{PaymanZ17,LiuJLA17,JuvekarVC18,MishraLSZP20,SafeNET,wagh2020falcon,wagh2019securenn,mohassel2018aby3}, \ie methods for collaborative computation without revealing the underlying data using primitives such as garbled circuits, oblivious transfer, and secret sharing. 

\para{Relationship to DM}
These techniques assume the setting where the Collector also plays the role of the Server (see~\cref{ssec:framework:actors}).
The data is generally encrypted on the Client (Pre-Hoc, can be seen as transformative DM), and sent in encrypted form to the Collector, who can then perform training or inference without ever seeing the data in the clear. 
The protection against \advtwire{} and \adviwire{} thus directly follows from the cryptographic guarantees of the underlying encryption schemes.
However, to use encryption, all discussed instantiations require the Client to have certain computational resources, and MPC-based approaches additionally require active Client participation.
This induces high computational and communication overhead---while recent research shows promise with strong guarantees, they are not yet practical for many (increasingly emerging) large-scale applications.

\smallskip\noindent\paragraph{DM Observations:}
\begin{itemize}[leftmargin=*]
    \item Scaling secure computation is directly aligned with vertical DM: simplifying/removing input features can noticeably reduce computational overhead of underlying algorithms.
    \item Under ideal FHE training, the final model is identical to the one trained on the original data, thus secure computation offers no protection against \advmodel{}.
    \item As the Collector only ever has access to encrypted data, \advstore{} is not a relevant concern in this case. 
    \item[\large \smallicon{\faLightbulbO}] The cryptographic guarantees offered by secure computation are often complementary to those of other techniques, raising the idea of combining to achieve more holistic DM. While some combinations were already studied (\eg federated learning~\cite{bonawitz2016practical}), many others are still unexplored, leaving an interesting avenue for future work.
\end{itemize}

\vspace{-0.5em}
\subsection{Privacy-preserving Data Publishing}
\label{subsec:techniques:ppdp}

Privacy-preserving data publishing (PPDP) sanitizes a micro-data table or an aggregate table before release \cite{Fung2010}. %
In our hospital example, this step occurs when the hospital wishes to disclose patient information, either row‑level records or summary tables, to an external party.
It typically involves a deterministic transformation aiming to keep analytical value while mitigating reidentification. The classical {\em micro-data release} techniques include \emph{$k$-anonymity} \cite{Sweeney2002}, which ensures that each quasi-identifier tuple (\eg age and ZIP code, see \cref{app:data}) is generalized or suppressed so that it appears in at least $k$ patient records. Its refinements, \emph{$\ell$-diversity} \cite{ldivers}, which requires that each equivalence class contains at least $\ell$ distinct sensitive-attribute values (\eg diagnosis codes), and \emph{$t$-closeness} \cite{tclose}, which bounds the distance between the diagnosis distribution inside each class and the overall (hospital) distribution, further strengthen respective privacy guarantees as we will highlight below. The parameters ($k$, $\ell$, and $t$) give regulators thresholds to audit when assessing compliance with data minimization.

For {\em tabular release},  the hospital might instead publish a contingency table of disease counts. 
Key PPDP methods include \emph{cell suppression} \cite{Cox1980}, which removes cells violating dominance or that appear less than $k$ times. There exists also non-deterministic versions; notably, \emph{swapping} \cite{DaleniusReiss1982} that randomly exchanges attribute values (or entire records) across individuals while preserving low-order marginals, and \emph{differential privacy} \cite{dwork2006calibrating} (\cref{subsec:techniques:dp}), that relies on noise addition to provide privacy guarantees on the sanitized tables.

All these transformations achieve {\em vertical minimization} by generalizing or deleting attributes or {\em horizontal minimization} by coarsening or permuting rows. Importantly, as data release techniques, they constitute the backbone of statistical disclosure control handbooks and national-statistics practice.

\para{Relationship to DM}
There are two important aspects to consider when applying PPDP techniques in practice. First, the sanitization is typically performed by a trusted Collector. This means that the techniques cannot protect against \advtwire{}, but can protect against all later adversaries. Second, the sanitization is typically performed Post-Hoc, requiring Clients to provide full-resolution data.
Importantly, PPDP provides a set of guarantees:  In $k$-anonymity the adversary re-identification risk is upper-bounded by $1/k$ \cite{Sweeney2002}, however, the attribute-disclosure risk can remain high if all $k$ records share the same sensitive value. This is precisely the gap closed by $\ell$-diversity \cite{ldivers}. $t$-closeness further limits what an adversary with arbitrary background knowledge can learn, yet the bound is distributional, not individual, and thus weaker than differential privacy \cite{tclose}.  
Swapping has been shown to resist record linkage as long as the swap rate is sufficiently large \cite{Gomatam2005}, but formal leakage bounds are heuristic unless the procedure is embedded in a probabilistic analysis \cite{ChristRadwayBellovin2022}. 
We note the we classified PPDP as DM-adjacent in an ML context for two main reasons: (i) it's original intention targets purely data releases (instead of ML training) and (ii) it pre-dates most current DM regulations.

\smallskip\noindent\paragraph{DM Observations:}
\begin{itemize}[leftmargin=*]
  \item Note that information loss grows with $k$, $\ell$, or $t$: generalization widens intervals, suppression inserts `NA's, and swapping perturbs correlation structure, so the \emph{utility-privacy} trade-offs is directly apparent with minimization.
  \item %
  Additionally, suppression, by focusing on small counts, may introduce \emph{bias} in the data, as it removes records that are not necessarily sensitive but rare. This can lead to downstream fairness issues, as it may disproportionately affect underrepresented groups \cite{Fioretto:aaai24,FHZ:ijcai22}. 
  \item[\large \smallicon{\faLightbulbO}]  
  While PPDP is traditionally applied offline, the \emph{outputs} of an already-trained model can themselves constitute a form of data publishing. Such releases can leak training-set membership (through membership attacks \cite{Shokri2017, Yeom2018} or reveal sensitive attributes via model-inversion \cite{ModelInv}. Understanding how DM considerations could be integrated at this stage remains an important open question.
\end{itemize}

\vspace{-0.5em}
\subsection{(Fair) Representation Learning}
\label{subsec:techniques:frl}

Fair Representation Learning (FRL) refers to a family of methods designed to transform data into representations that satisfy fairness constraints with respect to a sensitive attribute while preserving downstream utility~\cite{zemel2013learning, madras2018learning, jovanovic2023fare}. More formally the objective is to produce a representation $z=f(x)$ minimizing information about the sensitive attributes $s$, while retaining information relevant to target tasks $y$.
Importantly, the encoder $f: \mathbb{R}^d \rightarrow \mathbb{R}^d$ is learned on a set of full-resolution data points and (unlike in PPDP) generally non-interpretable.

\para{Relationship to DM}
By design, FRL performs transformative vertical data minimization, aiming to reduce the presence of a sensitive attribute in the released representation $z$, thus limiting the information that any downstream party can exploit for inference or discrimination. This mapping enforces that the shared representation contains only task-relevant information (measured by utility loss) and is statistically independent or minimally informative with respect to the sensitive attributes (measured by a fairness loss).
Under typical usage, FRL operates as a pre-processing step performed by a trusted Collector, who processes the original data and only distributes the representation $z$ to a downstream Server. Thus, FRL mainly targets \advtpreproc, \advipreproc, \advstore, and \advmodel, mitigating inference attacks on sensitive attributes from $z$. As minimization occurs usually Post-Hoc, FRL does not limit access for adversaries at the collection stages (\advtwire, \adviwire) unless combined with other techniques.

\smallskip\noindent\paragraph{DM Observations:}
 \begin{itemize}[leftmargin=*]
\item Fairness is commonly adversarially evaluated using the same reconstruction adversary as for DM (\cref{app:adversarial}). The key connection is that \emph{as long as a downstream classifier can be unfair w.r.t. attribute $s$, an adversary must be able to (partially) reconstruct $s$ from the minimized $z$}.
\item[\large \smallicon{\faLightbulbO}] Several FRL methods provide guarantees on the downstream unfairness exhibited by any classifier. This opens up possible extensions of these guarantees to multiple-valued sensitive attributes while maintaining performance, \eg by actively retaining utility loss through synthetic data.
 \end{itemize}

\vspace{-0.5em}
\subsection{Synthetic Data}
\label{subsec:techniques:syn_data}
\vspace{-0.5em}

Synthetic-data methods train a generative model $G_\theta$ on real records $\bm{x} \sim D$ and release samples $\tilde{\bm{x}} \sim G_\theta$ in place of, or in addition to, the original data. Early work relied on parametric simulators~\cite{Rubin1993} while more modern approaches use Generative Adversarial Networks~\cite{Xu2019}, Variational Auto Encoders~\cite{Gondara2018}, diffusion models \cite{Azizi2023}, or large language models \cite{YeLi2023}. Because no row of $D$ is disclosed verbatim, synthetic data promises vertical data minimization by reducing feature granularity and horizontal minimization by decoupling any one individual from the released corpus.

\para{Relationship to DM}
In the vanilla setting the generator is trained centrally, so non-minimized data reaches the Collector. The synthetic release therefore protects only against later adversaries (Adv. \advtpreproc{}, \advstore{}, and \advmodel{}).
However, formal privacy guarantees (in terms non-disclosure risk) are absent unless training is combined with a $(\varepsilon,\delta)$-differential privacy mechanism, such as DP-GAN \cite{Xie2018} and PATE-GAN \cite{Jordon2019}. 
In particular, an over-parametrized generator introduces \emph{memorization risks} enabling reconstruction and membership inference \cite{stadler2022synthetic, Carlini2023}.
This overfitting is accentuated when the training set is small, so the DM benefit can invert in low-data regimes.
Finally, as synthetic datasets are fully shareable, they enable downstream minimization; irrelevant samples can be filtered without revisiting the original data, making it complementary to distillation and feature-selection methods.

\vspace{-0.5em}
\subsection{Dataset Distillation}
\label{subsec:techniques:dd}
\vspace{-0.5em}

Dataset distillation~\cite{wang2018dataset, sucholutsky2021soft, zhao23cond} is a method that synthesizes a compact dataset (often orders of magnitude smaller), which, when used for training, aims to achieve comparable model performance to full dataset training. Unlike data selection \cite{nalepa2019selecting, mirzasoleiman2020coresets}, distillation thereby optimizes synthetic data directly so that gradient steps taken on the distilled data (ultimately) mimic those taken on the full dataset \cite{lei2023comprehensive, cazenavette2022dataset, zhao2020dataset}.

\para{Relationship to DM}
The key data minimization benefit of dataset distillation lies in its horizontal minimization: by releasing only a distilled (synthetic) subset, it reduces the volume of data exposed to downstream parties, thus protecting all examples omitted from the distillation. As with synthetic data, if the distillation is performed by a trusted collector, some protection is provided against adversaries \advstore{} and \advmodel{}, but not against \advtwire{}, since original data must still be collected centrally before distillation. Similarly it does not protect Client data at inference time (\adviwire{}) where original data points are required.
Importantly, even though distillation may limit horizontal leakage, unless explicit privacy constraints are incorporated (e.g., DP-Distillation~\cite{vinaroz2023differentially}), there is no formal privacy guarantee against adversarial inference from the distilled examples~\cite{carlini2022freelunchprivacyfree}, making DM-centric distillation (e.g., jointly  across both samples and features while incorporating a privacy term) an interesting area for future exploration ({\faLightbulbO}).

\vspace{-0.5em}
\subsection{Data Selection}
\label{subsec:techniques:ds}

Driven by data-efficiency, ML research has produced a range of algorithms reducing the amount of data required to train models to a given performance level. Early methods relied on \emph{coresets} \cite{mirzasoleiman2020coresets} of examples which encompass representative knowledge for the underlying task. More recent work on large-scale models \cite{kaddour2023minipile, li2024datacomp} has considered various methods such as data deduplication, quality filtering, as well as other types of performance and relevance filtering.

\para{Relationship to DM}
Reducing the amount of data required for training provides horizontal DM guarantees against Adversary \advtpreproc~ and \advmodel. However, because only a subset is removed, these guarantees are limited to non-kept samples, providing no DM benefit on remaining instances. Additionally, current methods for selection require collecting all examples in advance, providing no protection from \advtwire{}, and only providing protection from \advstore{} if the original database is replaced with the curated version.

\smallskip\noindent\paragraph{DM Observations:}
\begin{itemize}[leftmargin=*]
    \item While reducing data collection provides a horizontal DM benefit, it is currently not privacy-focused, as selection methods are driven by utility or training efficiency. A more granular exploration of DM guarantees of specific data selection techniques may be interesting future work.
    \item Reducing dataset sizes at a fixed training budget leads to increased repetition of included training examples, known to increase memorization risk \cite{carlini2023quantifying} (i.e., to \advmodel).
    \item[\large \smallicon{\faLightbulbO}] 
    While data selection is typically performed on static datasets, the problem of filtering vertical data at inference-time ML remains largely underexplored. The development of adaptive policies is especially attractive for DM as it complements traditional post‑hoc publishing by limiting \adviwire{} and \advipreproc{} adversaries. 
\end{itemize}

\vspace{-0.5em}
\subsection{Data Augmentation}
\label{subsec:techniques:d_aug}

Data augmentation refers to the process of expanding a dataset by creating modified versions of existing data points~\cite{shorten2019survey, feng2021survey}. Typical methods include transformations such as image rotations, flips, crops, noise injection, or text paraphrasing, with the goal of improving model generalization by synthetically increasing data diversity. While primarily designed to combat overfitting or to enhance performance in low-data regimes, data augmentation interacts with data minimization in nuanced ways.

\para{Relationship to DM}
 At first glance, data augmentation appears antithetical to data minimization: instead of reducing, it expands the amount of data processed and stored. In fact, data augmentation for pure utility improvement has been shown to increase privacy risk \cite{yu2021does}.
However, augmentation can indirectly support minimization goals. By enhancing model robustness, it enables learning effective models from smaller original datasets—potentially allowing for thecollection or retention of less real data (\ie horizontal DM). Further, augmentation techniques that focus not purely on utility (e.g., cropping) can, in fact, reduce risks of membership inference attacks (\advmodel{}) \cite{kaya2021does}, providing an interesting avenue for future work ({\faLightbulbO}). 
Yet, most augmentation methods operate post-hoc and are utility-driven, offering no direct guarantees against any prior adversaries in our framework.

\vspace{-0.25em}
\subsection{Feature Selection}
\label{subsec:techniques:f_sel}
\vspace{-0.25em}

Feature selection techniques aim to identify and retain only the most relevant features (attributes) from a dataset for a given task~\cite{aliferis2010local, yu2019multi, vergara2014review, estevez2009normalized}. Commonly, approaches filter (e.g., using mutual information \cite{battiti1994using}),  recursively eliminate \cite{guyon2002gene} or implicitly select features (e.g., Lasso \cite{tibshirani1996regression} or decision trees \cite{grabczewski2005feature}). While classically motivated by reducing model complexity, improving generalization, and mitigating overfitting, it is also naturally aligned with data minimization by removing irrelevant or redundant features.

\para{Relationship to DM}
Feature selection is inherently vertical data minimization: removing features from the dataset entirely reduces the amount of personal data being collected and processed. This confers privacy benefits for individuals whose data would otherwise be exposed. Typically, feature selection is performed as a pre-processing step at collection time. Consequently, protection is afforded against adversaries \advtpreproc, \advstore, and \advmodel—once dropped, removed features are unavailable for subsequent attacks. However, feature selection in practice generally operates on the basis of maximizing model utility rather than directly optimizing for privacy or minimization; thus, the DM benefit is indirect and dependent on how aggressively the feature set is pruned.
It is important to note that conventional feature selection assumes the collector has access to all raw data, offering no protection against adversaries at collection time (\advtwire, \adviwire). Additionally, unlike existing works in explicit vertical Data Minimization (\cref{subsec:techniques:explicit}), selected features are retained at full granularity, so while the number of features is reduced, the resolution of the remaining attributes is not.

\smallskip\noindent\paragraph{DM Observations:}
 \begin{itemize}[leftmargin=*]
  \item Feature selection can be seen as the most extreme form of vertical data minimization, where the generalization of an attribute is to remove it altogether \cite{staab24vdm}.
 \item Focusing on utility, feature selection typically lacks explicit mechanisms to trade off utility versus privacy.

\item[\large \smallicon{\faLightbulbO}] 
    Similarly as for data selection, feature selection is typically performed on static datasets. When considering inference-time ML and streaming analytics feature selection must operate in an online and \emph{personalized} way (\eg different individuals may require to suppress different attributes to maximize model utility), thus requiring a mix of data and feature selection with local and adaptive policies \cite{tran23}. %

\end{itemize}

\subsection{Active Learning}
\label{subsec:techniques:act_learn}
Active learning \cite{settles2009active, li2024survey} and active feature value acquisition \cite{saar2009active} methods consider a setting where labels or additional features are collected in online queries, growing the pool of data as needed. These methods are originally designed for settings where data collection is expensive, for example, due to a dependence on human annotation. Methods for active learning may rely on improving the inductive biases of the learning algorithm to be more label efficient (through techniques such as meta- \cite{zhu2022few} or semi-supervised learning \cite{sener2017active}), or improving the querying strategy \cite{gal2017deep, ash2019deep}.

\para{Relationship to DM}
Active learning and active feature value acquisition are pre-hoc methods---data is only collected after a query is specifically issued for it. This provides protection from all training-time adversaries (\advmodel, \advtwire, \advtpreproc, \advstore) for the subset of examples whose data is not collected. Active learning provides no protection to those examples whose data is queried. These methods provide some combination of horizontal and vertical DM: vertical, as only a subset of feature values are collected for all examples, and horizontal, as values are only collected for those examples which are queried.

\smallskip\noindent\paragraph{DM Observations:}
\begin{itemize}[leftmargin=*]
    \item An increased reliance on a subset of data as in active learning may lead to increased privacy risks of that set~\cite{carlini2023quantifying}. Active feature value acquisition may have the opposite effect - privacy risk is generally lower when fewer features per example are collected \cite{dwork2015robust}.
    \item The requirement to issue queries to labels or feature values provides an interpretable notion of data collection to those examples whose data are collected.
    \item[\large \smallicon{\faLightbulbO}] Driven by reducing feature acquisition costs, active learning treats features as either fully observed or unobserved. However, it can be extended to selecting both the feature and its \emph{granularity}, still allowing the acquisition of finer feature resolution later, if needed. This opens a promising research direction for DM-focused active learning.
\end{itemize}

\subsection{Language Model Agents and RAG}
\label{subsec:techniques:rag}
Language models acting as personal assistants is an emerging use case of machine learning, with many recent works pursuing different aspects of this setting. 
In particular, this is an example where DM decisions are made at inference time: what data does the agent ingest at each step, what data can it share with other parties~\cite{ghalebikesabi2024operationalizing, mireshghallah2023can} including other agents, and what data can it share publicly. 
While fully describing the ongoing developments on this topic is out of the scope of this work, one use case that particularly well exemplifies the interaction with DM is related to retrieval-augmented generation (RAG). 
RAG is a method for providing relevant context to a language model, originally designed for improving factuality of model responses \cite{gao2023retrieval}. 
However, RAG can also be applied in privacy-sensitive settings, by including personal information into the context of a model only when needed \cite{zeng2024good, nvidia_chat_with_rtx} (instead of during model training). 
This approach is used in a variety of threat models: protection against the LLM provider, protection against the output of the model leaking information, or protection against the model sharing unnecessary information with other parties, all of which can be seen from the perspective of DM.

\para{Relationship to DM}
These methods all operate at inference time as vertical DM guarantees, limiting the information sent to or shared with the language model. As a result, they aim to protect against all inference-time adversaries.

\smallskip\noindent\paragraph{DM Observations:}
\begin{itemize}[leftmargin=*]
    \item No methods in presented this set provide formal guarantees, and care must be taken to protect these systems in any adversarial environments \cite{bagdasarian2024airgapagent, debenedetti2025defeating, zou2024poisonedrag, chaudhari2024phantom}.

    \item [\smallicon{\faLightbulbO}]
    While RAG pipelines are often designed to maximize utility, the \emph{retrieved context} itself can act as a vector of disclosure when it includes user-specific or sensitive content. Developing methods that dynamically select the minimal sufficient context (conditioned on task, model confidence, and user preferences) remains an open challenge \cite{bagdasarian2024airgapagent}.

    \item [\smallicon{\faLightbulbO}]
    How can these minimization policies be made context-aware and norm-sensitive, for example by drawing on theories such as Contextual Integrity \cite{Nissenbaum2004}? Answering this question could enable language agents to align data-sharing decisions with implicit social expectations and evolving user intent.
\end{itemize}

\subsection{Model Compression}
\label{subsec:techniques:model_comp}

Compression techniques, such as pruning \cite{lin2020dynamic}, quantization \cite{frantar2022optimal}, knowledge distillation \cite{hinton2015distilling}, and low-rank adaptation \cite{hu2022lora}, aim to reduce the memory footprint and/or improve the inference efficiency of trained models~\cite{dantas2024comprehensive}. 

\para{Relationship to DM}
By constraining the model’s capacity, compression can force it to minimize information not relevant to the task, showing conceptual alignment with data minimization principles~\cite{gonon2023can,huang2020privacy}.
However, such untargeted post-hoc compression techniques, can result in noticeable performance degradation~\cite{frantar2022gptq,lin2024awq}. To address this, models are typically fine-tuned during or after compression using the original training data or additional datasets. While improving performance, this also increases the model’s data exposure, raising privacy concerns~\cite{yuan2022membership}.
Importantly, even in cases where model compression techniques could offer privacy benefits, they only target \advmodel{}, lacking formal guarantees. Moreover, as information is removed from the trained model, it lacks any user agency or visibility of the minimized data.

\subsection{Explicit Studies of DM for ML} 
\label{subsec:techniques:explicit}

Beyond the previously described DM-adjacent techniques, several recent works address data minimization directly motivated by compliance requirements. In practice, these methods often closely align with some DM-adjacent techniques, allowing a joint categorization (presented in \cref{app:table:explicit}).

\para{Vertical DM}
One of the earliest works specifically addressing \emph{vertical data minimization} (vDM), which aims to reduce the granularity of individual data points, is by \cite{goldsteen}. This approach applies concepts from data anonymization \cite{Sweeney2002} by training decision trees guided by anonymization metrics (\emph{Normalized Certainty Penalty}) and accuracy-driven pruning. Resulting decision nodes determine attribute-level generalization boundaries. This method effectively protects data privacy during inference (\adviwire{}) but initially requires full-granularity data, leaving it vulnerable to initial training data interception (\advtwire{}). In a similar direction, \cite{staab24vdm} proposes vertical data minimization through a joint Gini-Impurity objective inspired by fair representation learning~\cite{jovanovic2023fare}. This approach achieves an improved privacy-utility tradeoff with significantly reduced runtime, requiring fewer full-resolution records and reducing \advtwire{} risks.
Unlike the fixed generalizations applied uniformly across individuals in the above approaches, \cite{ganesh2024dataminimization} proposes personalized data minimization, selectively excluding attributes for each individual. The underlying (personalized) optimization assumes complete dataset access, therefore targeting adversaries \advtpreproc{} and \advipreproc{}. 
Similarly, \cite{tran23} introduces an inference-time vDM approach aimed at minimizing attribute exposure by fully in-/excluding attributes based on classifier certainty, thereby reducing the risk of adversaries intercepting inference data (\adviwire{}). This mirrors a conceptual change in which explicit DM methods, while building on adjacent techniques, \emph{increasingly shift focus to inference privacy}.

\para{Horizontal DM}
Next to approaches focusing on instance-level minimization, there's also work on \emph{horizontal} data minimization, i.e., directly reducing the number of data points collected. Even more so than in vDM, existing work in \emph{hDM} such as \cite{shanmu}, is closely related to DM-adjacent concerns such as active feature value acquisition as well as the reduction of overall storage and data collection cost \cite{mirzasoleiman2020coresets, paul} (\cref{sec:framework}). While hDM approaches do not (technically) reduce worst-case individual privacy risk (\cref{app:adversarial}), they are effective at lowering the population-level privacy risk and can, in practice, often be easily combined with vDM approaches. 
Importantly, hDM naturally relates to the regulatory notion of necessity \cref{app:legal_op}, in particular promoting approaches that aim to be data-efficient at reaching a targeted accuracy.

\para{Application Specificity}
Further, there exist application-specific data minimization studies such as \cite{biega2020operationalizing} that focus on DM in the \textit{recommender setting}. In particular, they compare neighborhood and matrix-factorization-based vDM for movie/location recommendations. While such approaches come with specific requirements (e.g., each user observed/rated a different set of movies), overarching DM principles stay the same. This includes reducing the initial training data size to learn minimizers, limiting the threat of \advtwire{} while primarily focusing on \adviwire{}. Recommender systems are also a particular focus of \cite{niu2023leveraging}, which use uncertainty-estimate inspired attribute or user-removal strategies, separately investigating vertical and horizontal data minimization. Lastly, \cite{anciaux2024new} focuses on a rule-based informed data minimization procedure, offering users (potentially) multiple choices when filling a form. Although primarily focusing on simpler decision-making processes (usually not requiring an ML pipeline), this approach minimizes data already on the client side, protecting against potential adversaries on the wire. At the same time, it only allows for full feature selection, which, depending on the setup, might overly expose selected attributes (\cref{app:data}).

In contrast to operational methods, other works primarily focus on auditing or contextualizing existing DM solutions. \cite{raste} introduces an attribute-level \emph{need-to-know} principle balancing privacy and fairness, while \cite{raste2} provides black-box auditing to assess DM based on model stability analyses.

%% file: tables/techniques.tex
\begin{table*}[!t]
    \renewcommand{\arraystretch}{0.8} %
    \centering
    \caption{Overview of DM-adjacent techniques across dimensions introduced in~\cref{ssec:framework:dimensions}. For each technique described in \cref{sec:techniques}, we list common instantiations, as well as the type of DM-centric data transformation they apply (\emph{type}). Further describe \emph{where} and \emph{when} in our workflow (\cref{sec:framework}) it is applied (encoding trust assumptions) and whether it differs between individual Clients (\emph{Personalized}). Lastly, we overview the targeted adversarial scenario and potential privacy guarantees.}
    \label{tab:overview_adjacent}
    \vspace{-1mm}

    \begingroup            %
      \renewcommand\citepunct{, \allowbreak\hspace{0pt}}%
      \setlength{\parindent}{0pt}

    \resizebox{\linewidth}{!}{
        \begin{tabular}{p{0.08\linewidth}p{0.14\linewidth}p{0.08\linewidth}p{0.08\linewidth}p{0.05\linewidth}p{0.05\linewidth}p{0.085\linewidth}p{0.1\linewidth}p{0.03\linewidth}}
            \toprule
            \textbf{Technique} & \textbf{Instantiation} & \textbf{Type of DM} & \textbf{Where?} & \textbf{When?} & \textbf{\shortstack[c]{Personal-\\ization}} & \textbf{\shortstack[c]{Client\\Requirements}} & \textbf{\shortstack[c]{Target\\Adversary}} & \textbf{\makebox[0pt][l]{\hspace*{-2.1em}\shortstack[c]{Privacy\\Guarantee}}} \\ \midrule
            Federated Learning & \cite{mcmahan2017communication, zhu2019deep, boenisch2023curious, bonawitz2016practical, kairouz2021advances}
 & \DM{T} & \PD{TC} & \PH{Pre} & \makecell[tl]{\PR{None}\\[1ex] \PR{Exp}} & \makecell[tl]{\CR{Act}\\[1ex] \CR{Tech}} & \AS{1} & \ding{55} \\
            \midrule
            \multirow{2}{*}{\parbox[t]{\linewidth}{Differential\\Privacy}} & Central \cite{dwork2006calibrating, abadi2016deep} & \DM{T} & \PD{TS} & \PH{Post} & \PR{None} & \CR{None} & \AS{6} & \ding{51} \\
            \cmidrule[0.25pt](lr){2-9}
             & Local/Distributed \cite{kasiviswanathan2011can, dwork2006calibrating, bittau2017prochlo} & \DM{Vp} & \makecell[tl]{\PD{TC}\\[1ex] \PD{IC}} & \PH{Pre} & \PR{Exp} & \CR{Tech} & \makecell[tl]{\AS{1}\\[1ex] \AS{3}} & \ding{51} \\
            \midrule
            ~ & FHE Inference~\cite{GiladDLLNW16,BadawiJLMJTNAC21,BrutzkusGE19,LeeKLCEDLLYKN21,LeeLLKKNC21,LuHHMQ20} & \DM{T} & \PD{IC} & \PH{Pre} & \PR{None} & \CR{Tech} & \AS{3} & \ding{51} \\
            \cmidrule[0.25pt](lr){2-9}
            Secure & FHE Training~\cite{kim2018secure,kim2018logistic,nandakumar19towards,lou2020glyph,mihara2020neural} & \DM{T} & \PD{TC} & \PH{Pre} & \PR{None} & \makecell[tl]{\CR{Act}\\[1ex] \CR{Tech}} & \AS{1} & \ding{51} \\
            \cmidrule[0.25pt](lr){2-9}
            Computation & MPC Inference~\cite{PaymanZ17,LiuJLA17,JuvekarVC18,MishraLSZP20,SafeNET} & \DM{T} & \PD{IC} & \PH{Pre} & \PR{None} & \makecell[tl]{\CR{Act}\\[1ex] \CR{Tech}} & \AS{3} & \ding{51} \\
            \cmidrule[0.25pt](lr){2-9}
            ~ & MPC Training~\cite{PaymanZ17,wagh2020falcon,wagh2019securenn,mohassel2018aby3} & \DM{T} & \PD{TC} & \PH{Pre} & \PR{None} & \makecell[tl]{\CR{Act}\\[1ex] \CR{Tech}} & \AS{1} & \ding{51} \\
            \midrule
            PPDP / DA & \cite{Fung2010,Sweeney2002,ldivers,tclose,Cox1980,DaleniusReiss1982,dwork2006calibrating,Sweeney2002,Gomatam2005,ChristRadwayBellovin2022} & \makecell[tl]{\DM{Vg}\\[1ex] \DM{Vp}} & \PD{TS} & \PH{Post} & \PR{None} & \CR{None} & \AS{2} &\ding{55} \\
            \midrule
            \parbox[t]{\linewidth}{Fair\\Representation\\Learning} & \cite{zemel2013learning,fairvae,censorAdv,madras2018learning,moyer2018invariant,mcnamara,balunovic2021fair,kim2022learning,shui2022fair,jovanovic2023fare} & \DM{T} & \PD{TS} & \PH{Post} & \PR{None} & \CR{None} & \AS{2} & \ding{55} \\
            \midrule
            Synthetic Data & \cite{Xu2019, Gondara2018, Azizi2023, YeLi2023, Xie2018, Jordon2019, stadler2022synthetic} & \DM{T} & \PD{TS} & \PH{Post} & \PR{None} & \CR{None} & \AS{NA} & \ding{55} \\
            \midrule
            \parbox[t]{\linewidth}{Dataset \\ Distillation} & \cite{wang2018dataset, sucholutsky2021soft, zhao23cond, lei2023comprehensive, cazenavette2022dataset, zhao2020dataset} & \DM{T} & \PD{TS} & \PH{Post} & \PR{Imp} & \CR{None} & \AS{NA} & \ding{55} \\
            \midrule
            \parbox[t]{\linewidth}{Data\\Selection} & \cite{mirzasoleiman2020coresets, kaddour2023minipile, li2024datacomp, gadre2023datacomp} & \DM{H} & \PD{TS} & \PH{Post} & \PR{None} & \CR{None} & \AS{NA} &\ding{55} \\
            \midrule
            \parbox[t]{\linewidth}{Data\\Augmentation} & \cite{shorten2019survey, feng2021survey,yu2021does,kaya2021does} & \DM{H} & \PD{TS} & \makecell[tl]{\PH{Pre}\\[1ex] \PH{Post}} & \CR{None} & \CR{None} & \AS{NA} &\ding{55} \\
            \midrule
            \parbox[t]{\linewidth}{Feature\\Selection} & \cite{aliferis2010local,yu2019multi,tople2020alleviating,hasan2022understanding,battiti1994using,tibshirani1996regression,guyon2002gene,vergara2014review,estevez2009normalized, grabczewski2005feature} & \DM{Vs} & \makecell[tl]{\PD{TS}\\[1ex] \PD{IS}\\[1ex] \PD{IC}} & \makecell[tl]{\PH{Pre}\\[1ex] \PH{Post}} & \PR{None} & \makecell[tl]{\CR{None}\\[1ex] \CR{Tech}} & \makecell[tl]{\AS{2}\\[1ex] \AS{3}\\[1ex] \AS{4}\\[1ex] \AS{5}} &\ding{55} \\
            \midrule
            \parbox[t]{\linewidth}{Active \\ Learning (AL)} & \cite{settles2009active, li2024survey, saar2009active, zhu2022few, sener2017active,gal2017deep, ash2019deep} & \makecell[tl]{\DM{H}\\[1ex] \DM{Vs}} & \makecell[tl]{\PD{TS}\\[1ex] \PD{IS}} & \PH{Pre} & \makecell[tl]{\PR{None}\\[1ex] \PR{Exp}} & \makecell[tl]{\CR{Act}\\[1ex] \CR{Tech}} & \AS{NA} &\ding{55} \\
            \midrule
            LLM Agents and RAG & \cite{zeng2024good, ghalebikesabi2024operationalizing, mireshghallah2023can, nvidia_chat_with_rtx} & \DM{Vp} & \PD{IC} & \PH{Pre} & \makecell[tl]{\PR{None}\\[1ex] \PR{Exp}} & \makecell[tl]{\CR{Act}\\[1ex] \CR{Tech}} & \makecell[tl]{\AS{3}\\[1ex] \AS{4}\\[1ex] \AS{5} \AS{6}} & \ding{55} \\
            \midrule
            \parbox[t]{\linewidth}{Compression,\\Pruning,\\Quantization} & \cite{lin2020dynamic, frantar2022optimal, hinton2015distilling, hu2022lora, dantas2024comprehensive, zhu2024survey} & \DM{T} & \PD{TS} & \PH{Post} & \PR{Imp} & \CR{None} & \AS{NA} &\ding{55} \\
            \bottomrule
        \end{tabular}
    }
\endgroup              %
\end{table*}

%% file: src/takeaways.tex
\vspace{-0.5em}
\section{Discussion and Takeaways} \label{sec:takeaways}

Our analysis of \dmml-adjacent works in \cref{sec:techniques}, its presentation in our unified framework, and our discussion of datasets in \cref{sec:datasets}, together lead to several key insights that can guide both researchers and practitioners in the field.

\para{There Are Many Flavors of DM}
As evidenced by our analysis of the dimensions of DM in \cref{sec:framework} but also by the variety of techniques in \cref{sec:techniques}, DM is a broad concept that can have many different technical instantiations. This becomes especially clear when looking at adversarial targets: Techniques such as differential privacy guarantee the reduction of personal information in a model, however, at a much later stage, thus only partly fulfilling regulatory requirements. 
A key component for practitioners is to be clear about which part of their workflow (Collection, Training, Inference) to minimize before selecting suitable techniques.
Importantly, different flavors of DM are often orthogonal, allowing for their combination including their DM advantages.

\para{Techniques Can Have Unintended Effects on DM}
As shown above, in a \dmml setup, a technique can have DM-relevant effects that are not immediately obvious nor implied by the key focus of the technique. 
Sometimes, the DM impact can even be an unintended consequence of the approach. 
For example, it is well known that differential privacy requires more training data to reach a given level of performance than non-DP training; but collecting more data is not DM-neutral!
Another example is the relationship between model quality and DM: while it is intuitive that a more powerful model could provide accurate inference with less data, and is thus DM-positive, it has also been shown that more powerful models can leak more information about the training data~\cite{carlini2023quantifying}, having a DM-negative effect.
Practitioners should be aware of these effects and consider the unique interactions of their use-case with different techniques, ensuring that limitations are well understood.
Our formalization in \cref{sec:framework} can support this by providing a structured overview of existing DM dimensions.

\para{DM Evaluation is Not Standardized}
Importantly, our analysis highlights that there is no standard way to evaluate DM(-adjacent) methods.
In particular, there is no consensus on the best way to measure utility, privacy, or fairness, nor how to evaluate the tradeoffs between metrics.
On top of that, in \cref{sec:datasets} we highlight that there is no standard dataset for evaluating DM methods, with many methods being adapted (and restricted) to particular domains. 
To make research more understandable, the community should work towards developing standard evaluation metrics, guidelines, and datasets that remain interpretable to non-experts.

\para{Challenges in Operationalizing Regulatory Principles}
As pointed out in \cref{sec:techniques}, regulations are often phrased in conceptually straightforward terms (e.g. ``limited to what is necessary'') but are technically challenging to instantiate (e.g., when potentially (significantly) less minimization produces (slightly) higher downstream accuracy). This highlights an opportunity for greater dialogue between the technical and regulatory communities. To facilitate this, the research community can contribute by developing more standardized benchmarks and evaluation metrics that align with regulatory concepts like ``necessity'' and ``proportionality.'' Similarly, it is beneficial when researchers provide clear guidance on the privacy-utility tradeoffs inherent in their methods, enabling practitioners to make more informed decisions and enabling regulators to ground regulation in what is technically feasible.

\para{Research Gaps} 
Much of Section~\ref{sec:techniques} discusses research areas whose connection to DM has not been discussed in prior work, and we suggest some future directions to integrate them into DM research. However, the directions mentioned are far from exhaustive; we hope that researchers continue to explore the their connection. One broad opportunity for research is to investigate gaps in Table~\ref{tab:overview_adjacent}, where a  DM-adjacent lacks a desirable property (e.g., being personalizable). Understanding when different techniques are compatible (or not!) may also help improve the coverage of DM techniques. 

%% file: src/conclusion.tex
\vspace{-0.25em}
\section{Conclusion} \label{sec:conclusion}

In this SoK, we addressed a critical gap between Data Minimization principles and their application in the AI/ML domain by providing a comprehensive \dmml framework and systematically categorizing explicit DM focused and various DM-adjacent techniques.
Using clearly defined dimensions, we, for the first time, provided a unified view of the specific relationship each technique has with DM, allowing us to extract a set of practical insights.
Our structured approach aims to empower practitioners and researchers to navigate the complex landscape of DM.
In particular, we hope practitioners are able to use our work to learn about the DM techniques, understand their technical implications and tradeoffs, ultimately choosing the best option for their use case.
For researchers, our work aims to bridge the gaps between different communities by offering a unified DM-centric perspective on seemingly disconnected topics.
We hope that establishing a common vocabulary and identifying patterns across DM-adjacent fields can inspire future research, leading to novel ways of operationalizing DM.
\clearpage

%% file: appendix.tex
\input{src/app_moreresults}

\input{src/app_utility}

\input{src/datasets}
\input{src/app_reb}

%% file: src/app_moreresults.tex
\input{tables/techniques_explicit.tex}

\section{Adversarial Settings} \label{app:adversarial}

We overview common metrics used to quantify the DM privacy risk, discussing their tradeoffs, limitations, and relationship to new concerns arising with generative AI.

\subsection{Reconstruction Attacks}

The most common DM-specific privacy metric is the \emph{reconstruction adversary}, which measures how effectively original sensitive attributes can be reconstructed from minimized data (see Staab et al., (2024) \cite{staab24vdm} (Section 5) for a formal introduction).
Commonly, such adversaries are instantiated as neural networks, serving as an estimate of the reconstruction success of the theoretical \emph{optimal adversary}.

\para{Gap to the Optimal Adversary} 
Such estimates are only lower bounds on true privacy risk under the optimal adversary, as the empirical adversary is limited by (i) the chosen model class (ii) optimization challenges that may lead to \eg local minima (iii) the availability of full-granularity training samples.
While theoretical bounds on this gap are desirable, current approaches rely on empirical approximations \cite{staab24vdm}, highlighting an important area for future work.

\para{Individual Risk vs. Population Risk} Another issue is that the optimal adversary generally aims to maximize \emph{population-level} accuracy, while existing regulations are generally phrased in terms of \emph{individuals}. Thus, even under low (optimal) adversarial risk reconstruction adversaries offer no guarantees that any individual is protected. %

\para{Impact of Side Information} Reconstruction attacks benefit from external \emph{side information}, i.e., additional data that the adversary might possess about the targeted individuals. 
This risk increase is particularly severe if the side information is highly correlated with sensitive attributes or overlaps with the minimized dataset. Practitioners must therefore consider this when assessing the privacy guarantees of DM.

\para{Relevance of Sensitive Attributes}
A further consideration when reporting reconstruction risk is the choice of aggregation across sensitive attributes. 
First, unlike most prior works in PPDP, DM targets all personal attributes, in part necessitating an aggregate metric over them.
Second, many attributes have varying baselines (e.g., random guessing a binary attribute achieves $\sim 50\%$ accuracy while guessing integer age is much harder), making adversarial reconstruction accuracies incomparable.
Is it better to have $80\%$ adversarial accuracy on attribute A and $20\%$ on attribute B or vice-versa? We advocate that evaluations, in addition to aggregate scores, always report their individual attribute reconstruction rates. 

\para{The Privacy-Utility Tradeoff}
This challenge in aggregating reconstruction risk across heterogeneous attributes touches on the broader privacy-utility tradeoff in (vertical) DM: Stronger generalization or aggressive feature suppression may successfully reduce adversarial success rates but often leads to diminished utility for the intended task. DM evaluation must balance these competing objectives, adding an additional dimension to the results---in practice, works often report the privacy-utility Pareto frontier across a set of minimizations \cite{staab24vdm}, allowing practitioners to control the tradeoff.

\subsection{Other Metrics}

\para{Linkability and Re-identification}

Linkability and reidentification attacks focus on associating minimized data with other datasets or external information sources, compromising individual privacy indirectly~\cite{dpwp}. These attacks commonly occur when adversaries leverage external side information to enhance their inference capabilities.
Typical use cases include evaluating scenarios involving multiple data releases or shared attribute structures across datasets. A significant limitation of linkability metrics is their dependence on available external datasets, potentially leading to underestimates if such datasets are unknown or underestimated.

\para{Singling Out}

Singling out attacks aim to uniquely identify an individual within minimized datasets~\cite{dpwp}. Such attacks exploit low-frequency generalizations, resulting in privacy risks associated with insufficient anonymity.
The effectiveness of singling out attacks is typically measured by the existence of unique or rare generalized records in the dataset. While this attack captures critical privacy risks, its effectiveness largely depends on dataset size, diversity, and granularity of generalization. Practitioners must thus carefully balance utility and granularity to prevent successful singling out.

\para{Membership Inference}
Membership inference attacks (MIAs) evaluate the privacy risks associated with DM by measuring whether an adversary can determine if an individual's data was used during model training. 
Conceptually, membership attacks leverage differences in model outputs to distinguish between training and non-training data points. 
Importantly, this relates more to horizontal DM (inclusion of full-resolution data) than vertical DM. 
As shown in \cref{subsec:techniques:dp}, Differential Privacy, while providing guarantees against MIAs, generally only targets \advmodel{}, failing to protect against other adversaries. 

\para{Memorization}
Memorization occurs when generative AI models unintentionally reproduce exact or near-exact training examples in their outputs.
It is prominent in LLMs, where it relates to both horizontal DM (similar to a MIA on text training data) and vertical DM (privacy issues arise primarily because individual data points contain personal information). 
Memorization is generally measured via \textbf{data extraction attacks}, where adversaries attempt to retrieve verbatim data by prompting the model with specific inputs \cite{carlini2021private}---this has been criticized in later work, which highlights how even approximate reconstructions \cite{ippolito2022preventing} constitute privacy risk.

%% file: tables/techniques_explicit.tex
\begin{table*}[t]
    \centering
    \caption{ Overview of explicit studies of \dmml (\cref{subsec:techniques:explicit}) from the perspective of the DM dimensions (\cref{ssec:framework:dimensions}).}
    \label{app:table:explicit}
    \vspace{-1mm}
    \resizebox{\linewidth}{!}{
        \begin{tabular}{p{0.2\linewidth}p{0.08\linewidth}p{0.08\linewidth}p{0.05\linewidth}p{0.05\linewidth}p{0.08\linewidth}p{0.1\linewidth}p{0.18\linewidth}}
            \toprule
            \textbf{Instantiation} &  \textbf{Type of DM} & \textbf{Where?} & \textbf{When?} & \textbf{\shortstack[c]{Personal-\\ization}} & \textbf{\shortstack[c]{Client\\Requirements}} & \textbf{\shortstack[c]{Target\\Adversary}} & \textbf{\shortstack[c]{Privacy\\Risk Model}} \\ 
            \midrule
            Biega et al., (2020) \cite{biega2020operationalizing} & \DM{H} \DM{Vs} & \PD{TS} & \PH{Post} & \PR{Exp} & \CR{None} & \AS{1} & No targeted model \\
            \midrule
            Shamugam et al., (2021) \cite{shanmu} & \DM{H} \DM{Vs} & \PD{TS} & \PH{Pre} & \PR{Exp} & \CR{Act} \CR{Tech} & \AS{NA} & No targeted model \\
            \midrule
            Goldsteen et al., (2022) \cite{goldsteen} &  \DM{Vg} & \PD{IC} & \PH{Pre} & \PR{None} & \CR{None} & \AS{NA} & No targeted model \\
            \midrule
            Tran et al., (2023) \cite{tran23} & \DM{Vs} & \PD{IC} & \PH{Pre} & \PR{Exp} & \CR{Act} \CR{Tech} & \AS{3} & No targeted model \\
            \midrule
            Staab et al., (2024) \cite{staab24vdm} & \DM{Vg} & \PD{TC} \PD{IC} & \PH{Post} \PH{Pre} & \PR{None} & \CR{None} & \AS{2} \AS{3} \AS{5} & Reconstruction, Linkability and Singling Out adversaries: No Guarantees \\
            \midrule
            Niue et al., (2023) \cite{niu2023leveraging} &  \DM{H} \DM{Vs} & \PD{TS} \PD{IS} & \PH{Post} & \PR{Exp} & \CR{None} & \AS{NA} & No targeted model \\
            \midrule
            Ganesh et al., (2024) \cite{ganesh2024dataminimization} & \DM{H} \DM{Vs} & \PD{TS} & \PH{Post} & \PR{Exp} & \CR{None} & \AS{2} \AS{5} & Reconstruction, Linkability and Singling Out adversaries: No Guarantees \\
            \midrule
            Anciaux et al., (2024) \cite{anciaux2024new} & \DM{Vg} & \PD{IC} & \PH{Pre} & \PR{Exp} & \CR{None} & \AS{3} & Reconstruction and Linkability/Reidentification attacks: k-anonimity guarantee \\
            \bottomrule
        \end{tabular}
    }
    \vspace{-1em}
\end{table*}

%% file: src/app_utility.tex
\section{Private And Sensitive Data}
\label{app:data}
\input{tables/datasets}
\para{Sensitive Attributes and Quasi Identifiers}
As introduced in \cref{sec:techniques}, frameworks such as PPDP's operate under the more restrictive assumption of clearly defined sensitive attributes. Typically, PPDP methodologies presume the existence of just one sensitive attribute within a dataset, whereas other attributes are classified as either unique identifiers (directly identity revealing, e.g., SSN), quasi-identifiers (not individually revealing but can enable re-identification when combined, e.g., birth dates, ZIP, gender), or non-sensitive.

However, practical data often deviates from this simplified scenario, containing multiple sensitive attributes (e.g., health status, income level, religious affiliations), each capable of harming individuals' privacy if revealed. Moreover, modern regulatory frameworks, including the General Data Protection Regulation (GDPR)~\cite{gdpr}, mandate stricter privacy measures, stipulating comprehensive protection of all personal attributes rather than a selective subset. Here, minimization requires a holistic reduction of all personal data (including quasi identifiers) based on necessity for a specific purpose. As a result, a narrow definition of sensitive attributes may not meet expectations under regulatory frameworks.
Importantly, such a mismatch between existing literature and legal requirements also extends to other techniques such as Differential Privacy, which provide guarantees on membership inference, but no inherent protection from a reconstruction adversary (\cref{app:adversarial}). Thus while simplifications are necessary for technical feasibility, they can overlook contextual risk mitigation requirements.
This gap between technical solutions and regulatory requirements becomes more unclear across different data modalities beyond classical (tabular) data, where boundaries between personal and non-personal data attributes blur.

%% file: tables/datasets.tex
\begin{table*}[!t]
    \centering
    \caption{Overview of datasets common in DM(-adjacent) literature, grouped by category. We highlight the popularity of each category across different techniques from \cref{sec:techniques}, and overview their size, data sensitivity, and cost.}
    \vspace{-1mm}
    \resizebox{\linewidth}{!}{
        \begin{tabular}{
                p{0.14\linewidth}
                p{0.22\linewidth}
                p{0.33\linewidth}
                p{0.09\linewidth}
                p{0.15\linewidth}
                p{0.05\linewidth}
            }
            \toprule
            \textbf{Category} & \textbf{Examples} & \textbf{Popular In} & \textbf{Volume/Size} & \textbf{Sensitivity} & \textbf{Cost} \\ 
            \midrule
            \parbox[t]{\linewidth}{Low-Resolution\\Natural Images} & \parbox[t]{\linewidth}{CIFAR-10/100\cite{krizhevsky2009learning}, SVHN~\cite{netzer2011reading}\\ MNIST~\cite{lecun2002gradient}, FEMNIST~\cite{caldas2018leaf}} & Federated Learning, Differential Privacy, Secure Computation, Synthetic Data, Dataset Distillation, Data Selection, Data Augmentation, Active Learning & $100k{-}1M$ & Contextually Sensitive & Low \\
            \midrule
            \parbox[t]{\linewidth}{High-Resolution\\Natural Images} & ImageNet~\cite{deng2009imagenet}, COCO~\cite{lin2014microsoft}, Places365~\cite{zhou2017places} & Synthetic Data, Data Selection, Data Augmentation, Model Compression & $\sim 1k$ & Contextually Sensitive & High \\
            \midrule
            Face Images & CelebA~\cite{liu2015deep}, UTKFace~\cite{zhifei2017cvpr}, FairFace~\cite{karkkainen2021fairface} & Differential Privacy, (Fair) Representation Learning, Synthetic Data & $\sim 100k$ & Biometric Data; Demographic Data & High \\
            \midrule
            Census Tabular & Adult~\cite{adult_2}, Folktables~\cite{ding2021retiring} & Differential Privacy, Secure Computation, Privacy-Preserving Data Publishing, (Fair) Representation Learning, Synthetic Data, Feature Selection, Active Learning, Explicit \dmml & $\sim 100k$ & Demographic Data & Low \\
            \midrule
            Financial \& Medical Tabular & LendingClub~\cite{wordsforthewise_lendingclub_2020}, Purchase 100~\cite{Shokri2017}, MIMIC-III~\cite{johnson2016mimic}, eICU~\cite{pollard2018eicu} & Differential Privacy, Secure Computation, Privacy-Preserving Data Publishing, Synthetic Data, Feature Selection, Active Learning, Explicit \dmml & $100k{-}1M$ & Personal Data; Health Data & High \\
            \midrule
            Classification Text & AG News~\cite{zhang2015character}, GLUE~\cite{wang2018glue} & Synthetic Data, Dataset Distillation, Data Augmentation, Active Learning, Model Compression & $100k{-}1M$ & Contextually Sensitive & Moderate \\
            \midrule
            Curated Text Corpus & Shakespeare~\cite{caldas2018leaf}, Sent140~\cite{caldas2018leaf} & Federated Learning, Differential Privacy & $100k{-}1M$ & Personal Data & Low \\
            \midrule
            \parbox[t]{\linewidth}{Web-scraped \\ Raw Text Corpus} & Common Crawl (CC)~\cite{commoncrawl}, C4~\cite{raffel2020exploring}, The Pile~\cite{gao2020pile} & Data Selection & $100M{-}1B$ & Contextually Sensitive, sometimes Personal Data~\cite{subramani2023detecting} & Low \\
            \midrule
            Graph/Social Network & Facebook Graph~\cite{leskovec2012learning}, Twitter Graph~\cite{leskovec2016snap}, Reddit Graph~\cite{hamilton2017inductive} & Privacy-Preserving Data Publishing & $100k{-}1M$ & Personal Data & Moderate \\
            \midrule
            Recommendation Data & MovieLens~\cite{harper2015movielens}, Google Local~\cite{yan2023personalized}, Amazon Reviews~\cite{hou2024bridging} & Explicit \dmml & $100M{-}1B$ & Personal Data & Moderate \\
            \bottomrule
        \end{tabular}
    }
    \label{tab:datasets}
    \vspace{-1em}
\end{table*}

%% file: src/datasets.tex
\section{DM-adjacent Datasets} \label{sec:datasets}

Developments in machine learning are equally driven by techniques as the data to which they are applied. In this section, we, therefore, give a structured overview of widely used DM-adjacent datasets organized by modality, domain, sensitivity, and cost (\cref{tab:datasets}). This is particularly relevant as it gives practical insights into the data and attributes on which DM-adjacent methods are evaluated. By highlighting patterns in dataset usage, we further highlight overlooked areas and suggest directions for future research in data minimization.

\para{Dimensions}
We analyze datasets along three key dimensions: volume, sensitivity, and cost. The \textit{volume/size} of a dataset is defined as the number of individual data points it contains. The \textit{sensitivity} categories account for DM regulations, ranging from contextually sensitive information and broadly defined personal data to strictly regulated data such as biometric identifiers and demographic attributes~\cite{icouk}. Finally, the overall dataset \textit{cost} encompasses everything from the cost of raw data acquisition and storage to the required annotation effort, ranging from low-cost, freely available web-scraped text corpora to expensive, expert-labeled medical datasets.

\para{Technique-specific Dataset Popularity}
Unsurprisingly, different techniques naturally gravitate toward specific categories of datasets. For instance, fair representation learning focuses largely on tabular datasets with demographic features (e.g., Adult \cite{adult_2}, Folktables \cite{ding2021retiring}) and face image datasets (e.g., FairFace \cite{karkkainen2021fairface}, UTKFace \cite{zhifei2017cvpr}), where protected attributes are clearly defined. Similarly, privacy-preserving data publishing has been primarily explored in structured domains like census records or social graphs, which align well with the capabilities of traditional anonymization techniques. In contrast, model compression and dataset distillation are usually applied to large-scale image and text datasets (e.g., ImageNet \cite{deng2009imagenet}, GLUE \cite{wang2018glue}) to reduce training or inference costs on overparameterized models—an objective less relevant to tabular data. While techniques like differential privacy and active learning are more expensive, reflecting their foundational roles in both data-efficient learning and privacy considerations across modalities, their implementations and impact still vary significantly across domains. 
This mismatch in goals and domain-specific applications increases the comparability gap between DM-adjacent techniques.

\para{Structured Data in Explicit Studies of DM}
Interestingly, we observe that explicit studies of data minimization techniques have also concentrated almost exclusively on tabular and recommendation datasets---a fact that can be attributed to their conceptual closeness to specific DM-adjacent methods like feature (vDM) and data selection (hDM).
This narrow focus risks missing broader insights into how data minimization applies to complex data types. For instance, image data requires the selective redaction or obfuscation of identifiable features, through measures such as face blurring and image obfuscation. In textual datasets, minimizing identifiable attributes is challenging due to the subtlety and diversity of linguistic contexts.
Extending existing techniques to such real-world domains is, therefore, likely to become an equally challenging and high-impact frontier for future DM research.

\para{Data Sensitivity and Cost}
While some dataset categories (especially tabular data) potentially contain (disclosed) personal information (e.g., census data), many more complex domain datasets often do not reflect the kinds of sensitive data encountered in the real world. For instance, datasets like ImageNet~\cite{deng2009imagenet} for high-resolution natural images or GLUE~\cite{wang2018glue} for text classification are popular and well-curated but not representative of typical data that may include personally identifiable or otherwise sensitive content. This can obscure the true risks associated with these dataset categories, creating a gap between research and applications.

Finally, we emphasize that cost is not always a barrier to collecting sensitive data. While some sensitive data types (e.g., biometric, financial, or health records) may be prohibitively expensive, much personal data can be collected at scale, e.g., via large-scale web scraping. This both challenges existing regulations and, in turn, increases the risk of unregulated use. Such emergent use cases put scalable compliance-guided data minimization across domains into the center of emerging privacy-focused research.

%% file: src/app_reb.tex
\section{Operationalizing legal principles} \label{app:legal_op}

The EUs GDPR Article 5.1c directly describes \textit{data minimization} as the ``adequate, relevant and limited to what is necessary'' collection of data. 
For an effective operationalization of the DM principle, it is therefore not only important describe a technical DMML pipeline (\cref{sec:framework}) but also to lend technical interpretability of the corresponding terms and how they map to individual techniques. 
Even further, DM regulations do not exist in isolation. GDPRs Article 5.1a for example prescribes data collection to be \textit{transparent} while Article 5.2 explicitly requires \textit{accountability}, establishing additional data(-collection)-specific specific requirements in a practical DMML workflow. 

In this section, we make a first effort of relating these regulatory notions to more specific properties of individual DM(-adjacent) techniques presented in \cref{sec:techniques} by linking these principles to explicit points in the pipeline and their corresponding data transformations. For the purpose of this we focus on three summarized principles consistently present across DM regulations: \textit{necessity},  \textit{proportionality}, and the aforementioned \textit{accountability}.

\paragraph{Necessity}
The principle of necessity requires that only the information strictly needed for a given purpose is collected, processed, or disclosed. Within the DMML pipeline, this can be interpreted as identifying the minimal sufficient features, or individual samples required to achieve an acceptable utility threshold for a given task.

Several techniques from \cref{sec:techniques} directly reflect this notion:
Feature selection and sparsification empirically determine which attributes are indispensable for downstream performance. Importantly, while proposed primarily from a performance perspective, common vertical approaches here are largely non-discriminatory (i.e., the same features are collected across all users) and may collect more data points than necessary.
Personalized vertical DM (e.g., active feature acquisition) improves on this by tailoring the required feature subset to the individual user or task context. Similar notions exist in inference-time DM (e.g., minimal-context retrieval in RAG), where approaches restrict exposed information to what is required for a specific query.
While, intuitively, more restrictive approaches seem strictly superior from a DM perspective, it is important to note that they often come at the expense of a higher operational burden (e.g., multiple interactions with a user instead of a single collection of features as in \cref{app:practical}).

Of separate interest here is horizontal DM, reflecting \textit{necessity} mostly from a training perspective. Approaches such as selective data acquisition, dataset pruning, and coreset construction aim to identify the smallest (i.e., \textit{necessary}) subsets of training examples to achieve certain performance.
This operationalizes necessity through explicit in- or exclusion of a data point. Importantly, such hDM approaches are generally not applicable during inference (i.e., the second half of our pipeline), where each data point is required for its corresponding prediction. As mentioned in \cref{sec:techniques}, we therefore view vDM and hDM as largely complementary methodologies in DM operationalization.

\paragraph{Proportionality}
The principle of proportionality requires that any additional data collection or privacy exposure is justified by a meaningful gain in utility. Within the DMML pipeline, this corresponds to tuning minimization parameters such that increases in information, granularity, or disclosure are balanced against their incremental risks.

Several techniques embody this idea. Differential privacy (DP) exposes privacy budgets (e.g., $\varepsilon$, $\delta$) that allow practitioners to select a point on the respective privacy-utility curve. PPDP approaches, such as $k$-anonymity or generalization hierarchies, let the data collector choose coarser or finer representations depending on the required level of protection. Sampling-based methods and active learning similarly quantify the marginal utility of collecting additional examples (drawing similarities to \textit{necessity}).
These trade-offs are only meaningful when interpreted with respect to the underlying adversarial targets and pipeline stage (\cref{ssec:framework:pipeline}). A proportional adjustment on the Collector side may, for instance, change the risk profile for Server-side adversaries. Our DMML framework makes these dependencies visible by linking such parameters directly to points of application and their respective threat models.

\paragraph{Accountability}
Accountability requires that minimization decisions can be justified and demonstrated in a transparent way. In DMML terms, this means documenting \emph{why} certain information is collected/modified/removed, \emph{how} these choices satisfy the aforementioned properties necessity and proportionality, and \emph{which} adversarial risks they address.

Our DMML framework supports this through its decomposition of minimization along three axes: (i) the minimization type (\textit{horizontal}, \textit{vertical}, \textit{transformative}), (ii) the point of application (Client, Collector, Server), and (iii) the relevant adversarial model (\advOne{}-\advSix{}). This structure enables practitioners to record for each technique---whether feature selection, hDM, PPDP, or DP---what information is restricted, why this restriction is appropriate, and how proportionality parameters were chosen. Importantly our pipeline for the first time provides a unified DM-centric view that allows a fair comparison between otherwise seemingly (dis-)connected techniques.

While accountability does not mandate a particular method, it provides a clear template for constructing auditable workflows that connect regulatory principles to concrete technical decisions. We hope that our framework can here aide in a more unified practical documentation and easier compliance descriptions in real-world DM settings.

\section{Practical Examples} \label{app:practical}

To complement the running example introduced in \cref{sec:framework}, we extend our hospital scenario with concrete illustrations of how different \dmml techniques may operate in practice. While the main paper analyzes each method more abstractly through the dimensions of our framework, we here aim to focus more on \emph{practical instantiations}: how specific techniques would be applied by Clients (e.g., Patients), the Hospital (Collector), or a Cloud-Provider (Server), which data flows they alter, and what advantages and trade-offs they introduce. These examples demonstrate how DM can manifest differently across horizontal, vertical, and transformative approaches, and how deployment choices influence trust assumptions, operational constraints, and privacy.

\paragraph{Federated Learning (\cref{subsec:techniques:federated})}
In our hospital scenario, federated learning allows the hospital to keep patient records on local devices or hospital-controlled edge systems while only sending model updates to the server provider. 
There are two varying typical instantiations one would consider here. Canonically, patients are local devices owners, only sending respective model updates to the Hospital.  
This realizes transformative, Client-side, Pre-Hoc DM for training data: raw records never leave the Client, and the cloud observes aggregated gradients instead of individual features. 
Another typical scenario extends the trust assumptions of the Client to his local hospital in which case FL is implemented with respect to an untrusted Meta-Collector (and Server) which trains a a model across multiple hospitals.
The main advantages are intuitive alignment with regulatory expectations (no central raw data collection) and reduced exposure to \advtpreproc{} and \advstore{}. However, FL requires Clients to perform local training and participate in communication, which increases technical requirements and system complexity. Moreover, model updates can still leak information about individual patients, so protection against \advmodel{} typically requires combining FL with DP or secure computation. However, if the clients or the hospital already own hardware capable of computing local updates (a strong assumption) later inference can also run locally, risking no potential exposure to or Server-side adversaries (\advFour{}).

\paragraph{Differential Privacy (\cref{subsec:techniques:dp})}
The hospital could also apply differential privacy when training models on patient data, either centrally or in combination with FL. In the central case, the hospital first collects all training data and then trains a model with DP noise added to the learning process, limiting what can be inferred about any single patient from the final model. This does not reduce the amount of data collected but provides formal guarantees against \advmodel{}. In a distributed or local variant, DP mechanisms are applied earlier, so Clients randomize their contributions before sending them to the hospital (and server), extending protection also to \advtwire{}, \advtpreproc{}, and \advstore{}. The key advantage is the availability of explicit, auditable privacy parameters; the main downside is the respective utility degradation (especially on lower privacy-budgets) and, in some cases, the need for more data to reach a given accuracy, which can be at odds with horizontal DM. Further, DP provides no guarantees during inference where clients want predictions on their actual data points.

\paragraph{Secure Computation (\cref{subsec:techniques:sec_comp})}
Secure computation offers an alternative in which a patient encrypts or secret-shares their data and lets the hospital and cloud provider perform training and inference directly on protected inputs. From a DMML perspective, this yields transformative, Pre-Hoc minimization: only encrypted representations leave the client, and the server never observes raw records, thereby mitigating \advtwire{} and \adviwire{}. This can be attractive when regulations or contracts prohibit the cloud from ever seeing plaintext data. The main drawbacks are practical: current secure-computation schemes impose substantial computational and communication overhead and require clients and the hospital to deploy specialized cryptographic infrastructure. Moreover, since the learned model is typically identical to the one trained on plaintext, there is no additional protection against \advmodel{} unless combined with other techniques.

\paragraph{Fair Representation Learning (\cref{subsec:techniques:frl})}
The hospital may also learn an intermediate representation of patient records before sending anything to the cloud. A fair representation model maps raw data to non-interpretable embeddings that aim to retain information relevant for disease prediction while suppressing dependence on sensitive attributes (e.g., ethnicity or insurance status). These embeddings are then shared with the cloud instead of the original features, realizing transformative, Collector-side DM that primarily targets \advtpreproc{}, \advstore{}, and \advmodel{}. This can reduce the risk of downstream discriminatory use and leakage of sensitive attributes from shared representations. However, training the encoder still requires full-resolution data at the hospital (no benefit for \advtwire{}), and the opacity of the learned representation may complicate clinical validation and explanations towards regulators or patients. Importantly, fair representations apriori do not provide any guarantees on client privacy potentially leading to a false sense of security.

\paragraph{Synthetic Data (\cref{subsec:techniques:syn_data})}
Instead of sending real records, the hospital could train a generative model on its internal database and share only synthetic patient data with the cloud for model development. This provides Post-Hoc, Collector-side DM: the Server never sees real rows, which reduces exposure to \advtpreproc{}, \advstore{}, and \advmodel{} for non-released individuals. Synthetic data is easy to redistribute and can be reused across projects, which is attractive for research collaborations. However, the hospital must first ingest full-resolution data to train the generator, and overfitting can lead to memorization, reintroducing membership or attribute-disclosure risks. Formal protection typically requires combining synthetic data with DP, adding further utility costs and design complexity. Importantly, similar to (or even more than) federated learning, synthetic data generation makes strong assumptions on the hospital, having both expertise and the means to train large-scale generative models and produce privacy-preserving datasets. Even further SDG requires a large initial full resolution dataset of patient records, which is atypical as hospital share records to create such datasets in the first place. Lastly, similar to DP, SDG does not provide any inference-time privacy-protection where real patient records are required.

\paragraph{Horizontal DM via Data Selection (\cref{subsec:techniques:ds})}
The hospital can also apply horizontal DM by carefully selecting which patient records to send to the cloud, for instance using coreset methods or filters. Only a subset of patients then contributes to the outsourced training set, reducing the overall population exposed to \advtpreproc{}, \advstore{}, and \advmodel{}. This can significantly lower storage and processing costs while maintaining adequate predictive performance. At the same time, data selection usually operates on a fully collected dataset at the hospital, offering no protection from \advtwire{}, and it does not help individuals whose records are retained in the externally shared data. Concentrating training on a smaller subset may additionally increase memorization risk for included patients, highlighting a non-trivial trade-off between efficiency, DM, and individual risk. As already highlight in \cref{subsec:techniques:explicit}, hDM also does not provide any guarantees during inference.

\paragraph{Personalized Vertical DM and Active Feature Acquisition (\cref{subsec:techniques:act_learn,subsec:techniques:explicit})}
Beyond deciding which records to keep, the hospital may limit which attributes are collected for each patient. Personalized vertical DM and active feature acquisition implement this by initially asking only for a small set of core features (e.g., coarse demographics, a few lab values) and requesting additional attributes only when model uncertainty remains high. This yields fine-grained, Pre-Hoc vertical DM at the Client, directly reducing the volume and granularity of sensitive information that ever leaves the patient while still targeting a desired accuracy level. The downside is increased interaction and operational complexity: clinicians or digital intake systems must handle follow-up queries, and poorly chosen acquisition policies may either request excessive data (weakening DM) or too little (harming clinical utility/safety).